\documentclass[journal]{IEEEtran}
\usepackage{graphicx}
\usepackage{multirow}
\usepackage{epstopdf}
\usepackage[numbers]{natbib}
\usepackage{times}
\usepackage{amsmath}
\usepackage{amssymb}
\usepackage{amsfonts}
\usepackage{subfigure}
\usepackage[colorlinks]{hyperref}
\usepackage{xcolor}
\usepackage{colortbl}
\usepackage[ruled, linesnumbered, vlined]{algorithm2e}
\usepackage{algorithmic}
\usepackage{rotating}
\usepackage{adjustbox,lipsum}
\usepackage{dsfont}
\usepackage{booktabs}
\usepackage{multirow}
\usepackage{array}
\usepackage{makecell}
\usepackage{pifont}
\usepackage{listings}

\newcommand{\eg}{{\em e.g.}}          
\newcommand{\etc}{{\em etc}}

\begin{document}
	
	\title{BatStyler: Advancing Multi-category Style Generation for Source-free Domain Generalization}

	\author{Xiusheng Xu,
		Lei Qi, 
		Jingyang Zhou, 
		Xin Geng
		\thanks{The work is supported by NSFC Program (Grants No. 62206052, 62125602, 62076063), China Postdoctoral Science Foundation (Grants No. 2024M750424), Supported by the Postdoctoral Fellowship Program of CPSF (Grant No. GZC20240252), and Jiangsu Funding Program for Excellent Postdoctoral Talent (Grant No. 2024ZB242).}
		\thanks{Xiusheng Xu, Lei Qi, Jingyang Zhou and Xin Geng are with the School of Computer Science and Engineering, Southeast University, and Key Laboratory of New Generation Artificial Intelligence Technology and Its Interdisciplinary Applications (Southeast University), Ministry of Education, China, 211189 (e-mail: xuxiusheng@seu.edu.cn; qilei@seu.edu.cn; zhoujingyang@seu.edu.cn; xgeng@seu.edu.cn).}
		\thanks{Corresponding author: Lei Qi.}
	}
	
	\markboth{}%
	{Shell \MakeLowercase{\textit{et al.}}: Bare Demo of IEEEtran.cls for IEEE Journals}
	
	\maketitle
	
	\begin{abstract}
		Source-Free Domain Generalization (SFDG) aims to develop a model that performs on unseen domains without relying on any source domains. However, the implementation remains constrained due to the unavailability of training data. Research on SFDG focus on knowledge transfer of multi-modal models and style synthesis based on joint space of multiple modalities, thus eliminating the dependency on source domain images. However, existing works primarily work for multi-domain and less-category configuration, but performance on multi-domain and multi-category configuration is relatively poor. In addition, the efficiency of style synthesis also deteriorates in multi-category scenarios. How to efficiently synthesize sufficiently diverse data and apply it to multi-category configuration is a direction with greater practical value. In this paper, we propose a method called BatStyler, which is utilized to improve the capability of style synthesis in multi-category scenarios. BatStyler consists of two modules: Coarse Semantic Generation and Uniform Style Generation modules. The Coarse Semantic Generation module extracts coarse-grained semantics to prevent the compression of space for style diversity learning in multi-category configuration, while the Uniform Style Generation module provides a template of styles that are uniformly distributed in space and implements parallel training. Extensive experiments demonstrate
		that our method exhibits comparable performance on less-category datasets, while surpassing state-of-the-art methods on multi-category datasets. Code is available \href{https://github.com/Xuxiusheng/BatStyler}{here}. 
	\end{abstract}
	
	\begin{IEEEkeywords}
		Computer Vision, Source-Free Domain Generalization, Multi-Modal Models.
	\end{IEEEkeywords}
	
	%
	\IEEEpeerreviewmaketitle

	\section{Introduction}
	\IEEEPARstart{D}{eep} neural networks (DNNs) \cite{zhu2024solid, BalakrishnanZSG19, he2016deep, krizhevsky2012imagenet, simonyan2014very, Peng23} have achieved significant progress for various tasks, such as image classification \cite{huang2017densely, wang2017residual}, object detection \cite{carion2020end, faster2015towards}, and other visual tasks. However, it is well-known that DNNs tend to perform poorly on out-of-distribution test data \cite{hoffman2013one}. To tackle this problem, Domain Adaption (DA) techniques \cite{lee2022fifo, saito2019semi} have been extensively explored to learn domain-invariant features using source and unlabeled target data. In practical scenarios, obtaining unlabeled target data is not always feasible yet. Domain Generalization (DG) \cite{LiuXLTZ23, JinLZC22, QiWSG23, ChenLLT23} is developed to mitigate this limitation, it aims to improve the model’s generalization ability by using one or more source domains. However, due to the arbitrary nature of domain shifts, limited source domains will easily lead to model overfitting. Furthermore, it is costly and infeasible to collect and annotate multiple source domains data for training.
	
	\begin{figure}[t]
		\centering
		\subfigure[PromptStyler \cite{Cho_2023_ICCV}]{
			\includegraphics[width=4.12cm]{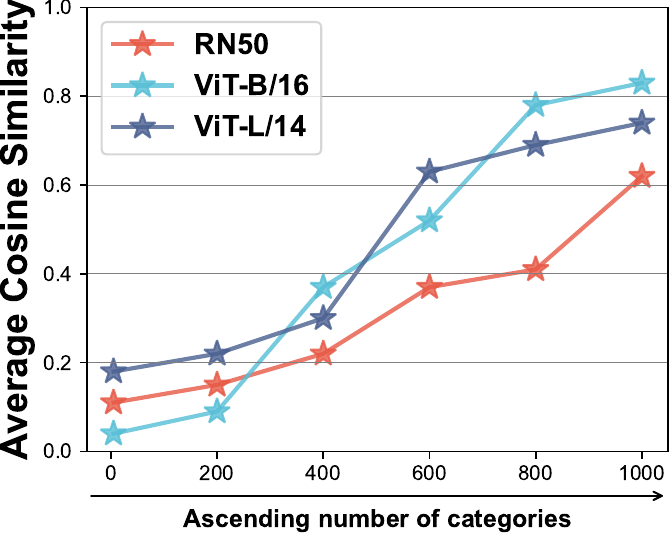}
			\label{figure:ps-cos}
		}
		\subfigure[BatStyler]{
			\includegraphics[width=4.12cm]{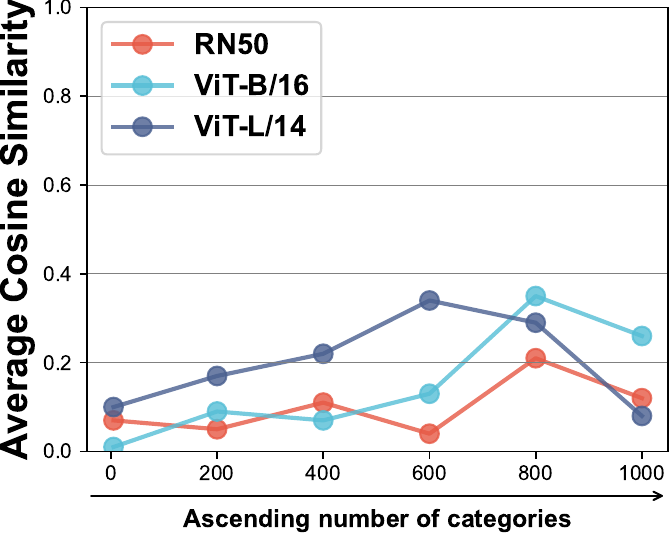}
			\label{figure:bs-cos}
		}
		\caption{\textbf{Average cosine similarity ($\downarrow$) of synthetic styles}. Comparison of PromptStyler and BatStyler on three models: ResNet-50, ViT-B/16 and ViT-L/14. We randomly sample 5, 200, 400, 600, 800 and 1000 category names from ImageNet-S. The number of style words is 80 and text features are obtained from prompt (\eg ~``a \textbf{\textit{S}} style of a") through text encoder.}
		\label{fig:1}
	\end{figure}

	\begin{figure}
		\centering
		\subfigure[PromptStyler \cite{Cho_2023_ICCV}]{
			\includegraphics[width=4.12cm]{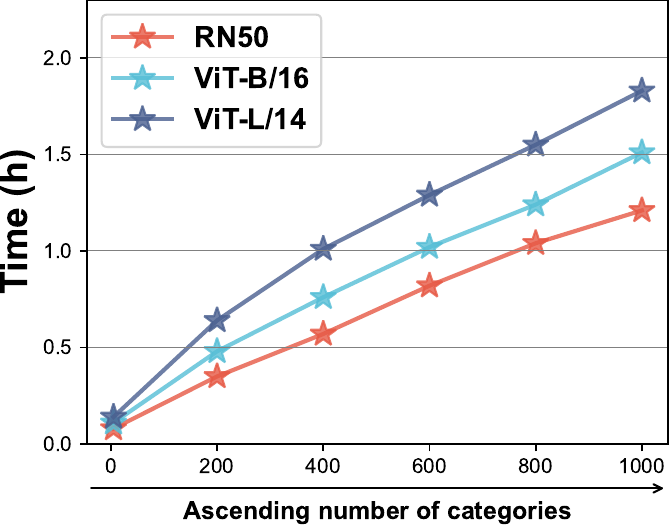}
			\label{figure:ps-time}
		}
		\subfigure[BatStyler]{
			\includegraphics[width=4.12cm]{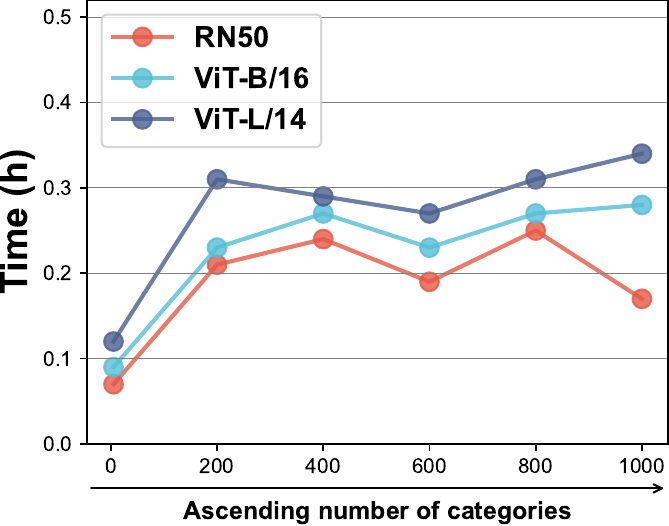}
			\label{figure:bs-time}
		}
		\caption{\textbf{Training time ($\downarrow$) of first training stage}. The experimental configuration adheres to the identical setup as described in Fig. \ref{fig:1}}
		\label{fig:2}
		\vspace{-0.3cm}
	\end{figure}
	Limited by the scale and diversity of DG datasets, it is difficult for existing methods to generalize to arbitrary domains in open-world scenarios. Therefore, Source-Free Domain Generalization (SFDG) \cite{0002CKRT22, niu2022domain, KunduKSJB21} task is necessary and challenging in real world scenarios. It assumes no access to images during the training and relies solely on the definition of target task, including category names. Thanks to the research of vision-language models, existing SFDG methods typically adapt a pre-trained model to target domains \cite{fang2022source, huang2021model, 0001S0Z24}, but these methods suffers from both unsatisfactory accuracy due to limited use of target domains.
	
	Recently, Cho \textit{et al.} \cite{Cho_2023_ICCV} propose a prompt-driven style generation method called PromptStyler, which utilizes a textual prompt to represent the corresponding image feature within a joint space of vision-language models (\eg ~CLIP \cite{radford2021learning}), thus eliminating the dependence on images. Nonetheless, there still exist two issues: \textbf{1)} PromptStyler ensures semantic consistency through all downstream category names. As shown in Fig. \ref{figure:ps-cos}, with the number of categories increases, semantic consistency will hinder the learning of style diversity. The cosine similarity between styles are continuously rising, which means that increasingly poorer diversity. \textbf{2)} The one-by-one training strategy is too slow, limiting the training efficiency when the number of styles and category names are very large, as shown in Fig. \ref{figure:ps-time}. To address these two issues, we propose BatStyler to further unlock the space of diversity and provide styles that are more uniformly distributed. Leveraging the characteristic of uniform distribution, we achieve parallel training of styles.
	\vspace{-0.4cm}

	In this paper, we introduce a novel method: \textbf{BatStyler}, which is utilized to enhance the style diversity in multi-category scenarios. We design our framework from the following two perspectives: 1) \textit{How to weaken the overly strong constraints of semantic consistency on style diversity}? Experimental results and theoretical evidence reveal that the semantic consistency compress the space of style diversity, which leads to the obstruction in diversity learning, especially in the case of a large number of categories. The core issue is to weaken the overly strong constraints of semantic consistency while retaining sufficient semantics to prevent styles from learning completely at random, we model this issue as an optimization problem, with the objective of maximizing style diversity and the regularization term being semantic consistency. For more details, please refer to Sec. \ref{sec:csg}. We consider that when the number of categories is very large, some categories have a high similarity (\eg ~`tabby cat' and `tiger cat'). Therefore, we filter out redundant fine-grained semantic constraints to enhance style diversity. Before training, we extract coarse-grained semantics from all categories in task through LLM, resulting in \textbf{C}oarse \textbf{S}emantic \textbf{G}eneration \textbf{(CSG)}. 2) \textit{How to generate style uniformly distributed in joint space}? In \cite{Cho_2023_ICCV}, style diversity is trained by orthogonality, which means there must be a large portion of entire space that remains uncovered (\eg ~80 orthogonal vectors can not cover a 1024-dimension space according to schmidt orthogonalization). To address the deficiency, we propose a module called \textbf{U}niform \textbf{S}tyle \textbf{G}eneration \textbf{(USG)} to initialize a set of uniformly distributed vector templates through neural collapse, ensuring that the generated styles are evenly distributed throughout the entire space. We present this set of templates in the form of a classifier, which conveniently enables parallel training of styles, thereby improving training efficiency.
	
	In summary, our main contributions can be listed as:
	\begin{itemize}
		\item We propose an innovative approach for SFDG in multi-category scenarios that enhance the style diversity of synthetic data by weakening the strength of semantic constraints and providing a  style template that uniformly distributed in the joint space. 
		\item We integrate style templates into a fixed classifier to achieve parallel training of styles. 
		\item Extensive experiments demonstrate the comparable performance on less-category datasets and superior performance on multi-category datasets compared with state-of-the-art methods.
	\end{itemize}
	
	The structure of this paper is outlined as follows: Section \ref{section:related work} provides a literature review on relevant research. In Section \ref{section:method}, we introduce our proposed method: BatStyler. Section \ref{sec:experiment} presents the experimental results and analysis. Lastly, we summarize this paper in Section \ref{section:conclusion}.
	
	\section{Related work}
	\label{section:related work}
	In this section, we will review the related approaches to our method, including domain generalization, vision-language models and neural collapse. The following part presents a detailed investigation.
	\vspace{-0.2cm}
	
	\begin{figure*}[t]
		\centering
		\includegraphics[width=18cm]{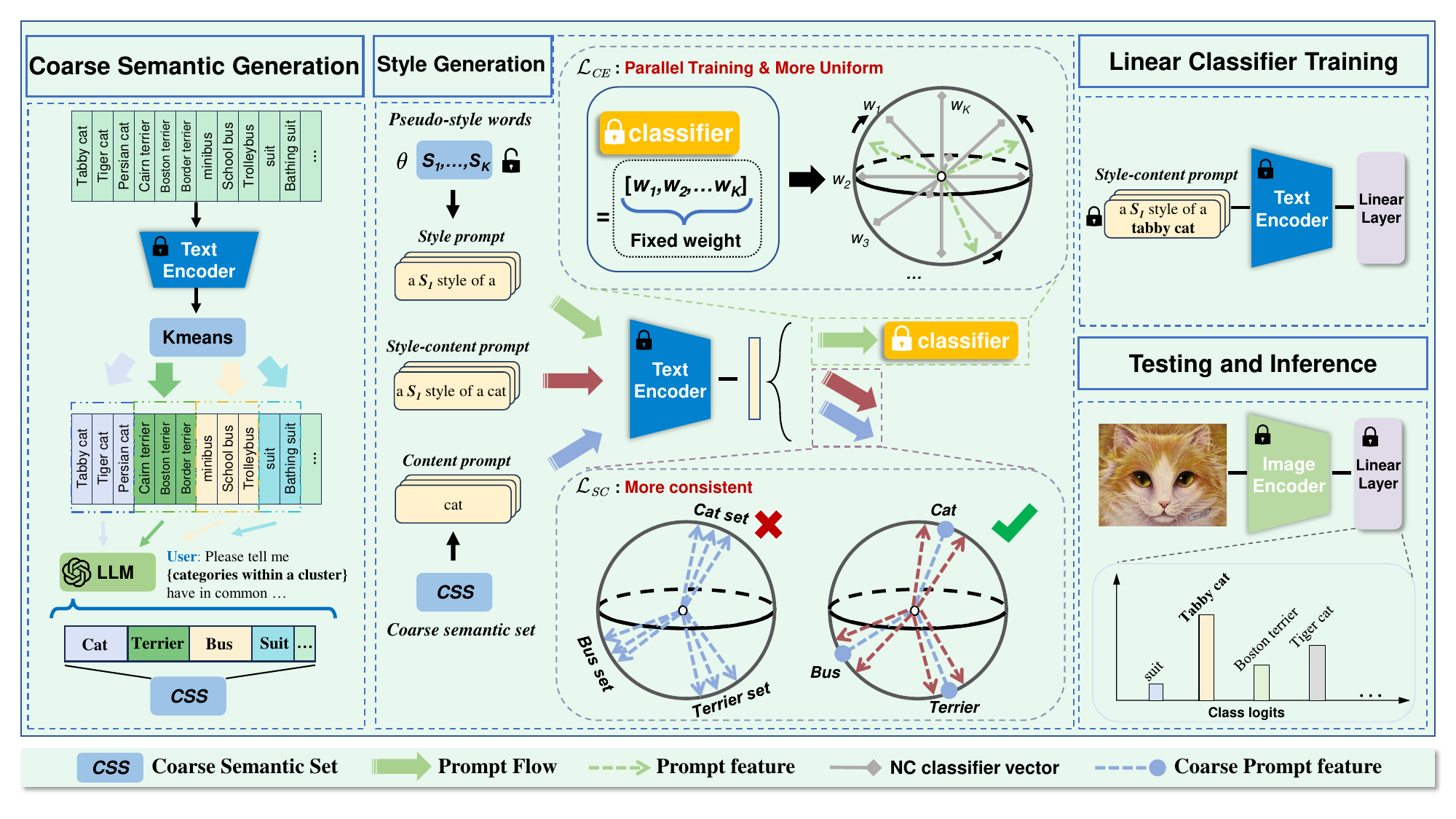}
		\caption{Overview of BatStyler. A Coarse Semantic Generation module is used to extract coarse-grained semantics of downstream categories. A classifier initialized by neural collapse is used to perform parallel training and generate styles that are more uniform distributed, which produces a better style diversity and higher training efficiency. For semantic consistency, we employ the extracted \textbf{C}oarse-grained \textbf{S}emantic \textbf{S}et (\textbf{\textit {CSS}}) to ensure semantic consistency.}  
		\label{fig:pipeline}
	\end{figure*}
	
	\subsection{Domain Generalization}
	There has been a surge of interest in developing methods to tackle domain generalization problem in recent years \cite{FanWKYGZ21, QiaoZP20, ZhangMLHT18, FangOLY23}. Prior works aim to learn domain-invariant features by data augmentation \cite{NamLPYY21, NurielBW21}, feature alignment across domains \cite{zhou2020domain, PengBXHSW19} and feature normalization \cite{ZhaoXCCLZXKPSXY18, PanLST18, JiaRH19}. However, most of DG methods expect labelled data from one or more source domains to prevent overfitting, which limits the applicability to important application areas (\eg ~autonomous), where labelled data is scarce. Limited by the cost of collection and annotation, existing methods is difficult to apply to the real world scenarios. Therefore, source-free domain generalization is necessary and urgent.  
	
	Unlike traditional DG, source-free domain generalization \cite{Cho_2023_ICCV, niu2022domain} has no access to any images during training. Thus, SFDG is much more challenging than previous DG routes, but more in line with real-world applications. To realize this, previous works mostly achieve generalization through data augmentation via synthetic data \cite{ChenZSYWTY23, ZhaoXKLZWPSYF17, ZhaoCC00LLYF19}. Recently, Cho \textit{et.al} \cite{Cho_2023_ICCV} propose a prompt-driven style generation method called PromptStyler, which uses textual features to represent image features in a joint space. Specifically, the method preset a prompt (\eg ~``a \textbf{\textit{S}} style of a"), where the \textbf{\textit{S}} is a pseudo-word that can be trainable. Using style diversity loss to ensure that all encoded style features are perpendicular to each other and content consistency loss to put complete prompt (\eg ~``a \textbf{\textit{S}} style of a [class]") and category name (\eg ~``[class]") together to be parallel. Then, a linear classifier is trained using text features and inferred using image features. However, its style diversity and training speed will decline with the increase of style vectors and category names. We improve PromptStyler by introducing a coarse semantic generation module to weaken the constraints on style diversity and providing a style template that uniformly distributed in joint space, which enhance the style diversity in multi-category scenarios. 
	\vspace{-0.2cm}
	
	\subsection{Vision-Language models}
	Vision-Language models \cite{JiaYXCPPLSLD21, 0001LXH22, radford2021learning, zhu2023minigpt, ZhengWWQB22} play a key role in incredible strides of computer vision. Most visual tasks rely heavily on a classifier head in DNNs, and only can be applied to sole task, leading to poor transferability and lack of flexibility. To tackle these problems, vision-language models (VLMs) are investigated extensively, which use image-text pairs as training samples and text as supervision, instead of hand-crafted labels. They can learn rich vision-language correlation from web-scale datasets and can be applied to other visual tasks without fine-tuning, including image recognition, captioning, retrieval and \etc. For instance, CLIP \cite{radford2021learning} presents a scalable contrastive pre-trained approach for learning a joint image-language space. With a vast corpus of 400 million image-text pairs, CLIP achieves superior zero-shot capabilities in visual tasks. Only modify the category names, it can migrate to other tasks. With the joint space of images and language, the two modalities are well migrated \cite{ZhangHHW0Y23}. Based on this joint space, \cite{Cho_2023_ICCV} employs textual features to train a linear classifier, and then the classifier is directly transferred to the image encoder to achieve image classification. In this paper, we primarily focus on its poor style diversity in multi-category scenarios to improve performance, thereby enhancing the generalization capability. 
	
	\subsection{Neural Collapse}
	Recently, a new phenomenon called Neural Collapse (NC) \cite{papyan2020prevalence} has been discovered, which describe an elegant geometric alignment between last-layer features and classifier in a well-trained model. DNNs enter the terminal phase of training and tend to exhibit intriguing neural collapse properties when training error reaches zero. NC essentially clarifies a state at which the within-class variability of last-layer outputs is infinitesimally and their class means form a simplex Equiangular Tight Frame (ETF), which refers to a matrix that composed of N maximally-equiangular and equal-$\ell_2$ norm P-dimension vectors. \textbf{E}=[$e_1$,...,$e_N$] $\in \mathbb{R}^{P \times N}$ and satisfies formula $\mathbf{E} = \sqrt{\frac{N}{N-1}} \mathbf{U} (\mathbf{I_N} - \frac{1}{N} \mathbf{1_N1_N^T})$, where \textbf{I}$_N \in \mathbb{R}^{N \times N}$ is the identity matrix, \textbf{1}$_N \in \mathbb{R}^N$ is an all-ones vector, and \textbf{U}$\in \mathbb{R}^{P \times N}$ is a partial orthogonal matrix that allows a rotation and satisfies $\mathbf{U}^T\mathbf{U}=\mathbf{I}_N$. All vectors in \textbf{E} have equal $\ell_2$-norm 1 and equal-maximal equiangular angle $-\frac{1}{N-1}$ \cite{YangCLXLT22}. 
	
	Although this theory is relatively new, there has already been a lot of works using NC in imbalanced learning \cite{XieYCH23, YangCLXLT22}, weakly-supervised learning \cite{XiaoFTZLCW24, HuWNTN24}, class-incremental learning \cite{RanLLTNT24, YangYLLTT23}, VLM tuning \cite{Zhu0ZYLKW24, zhu2023bridging} and other visual tasks. To address the issue of non-uniform style distribution, we initialize style vector templates for pseudo-styles uniformly distributed in joint space through neural collapse. Then we further adjust them using semantic consistency loss. At the same time, since it initializes pseudo-styles with largest margins between them all at once, we are able to achieve parallel training, improving the training efficiency significantly. 
	\section{The proposed method}
	\label{section:method}
	
	In this section, we will present our proposed framework BatStyler, which consists of two encoders, including a text encoder and an image encoder, a fixed classifier used to provide template for pseudo-styles, an coarse semantic generation module to extract coarse-grained categories and a learnable linear classifier used to classify images. Note that there is no correlation between the two classifiers mentioned above. 
	We employ CLIP as vision-language model, all parameters in CLIP are frozen in entire framework. The overview of proposed method is displayed in Fig. \ref{fig:pipeline}. The training process is divided into two stages: style generation and linear classifier training. The \textit{\textbf{Pseudo-style words}} is the only learnable parameter in style generation, which matches the $\theta$. The $\mathbf{\mathcal{L}_{CE}}$ is used to constrain style diversity and produce style features uniformly distributed in joint space based on style templates generated by neural collapse. The $\mathbf{\mathcal{L}_{SC}}$ is used to constrain semantic consistency. Due to the redundancy of semantics in multi-category scenarios, it compresses the space for style diversity learning, which motivates us to leverage the coarse-grained semantics to expand the random space for style diversity. We focus on style generation to improve poor diversity issues present in multi-category scenarios.
	
	\subsection{Coarse Semantic Generation}
	\label{sec:csg}
	
	\begin{figure}[t]
		\centering
		\includegraphics[width=8cm]{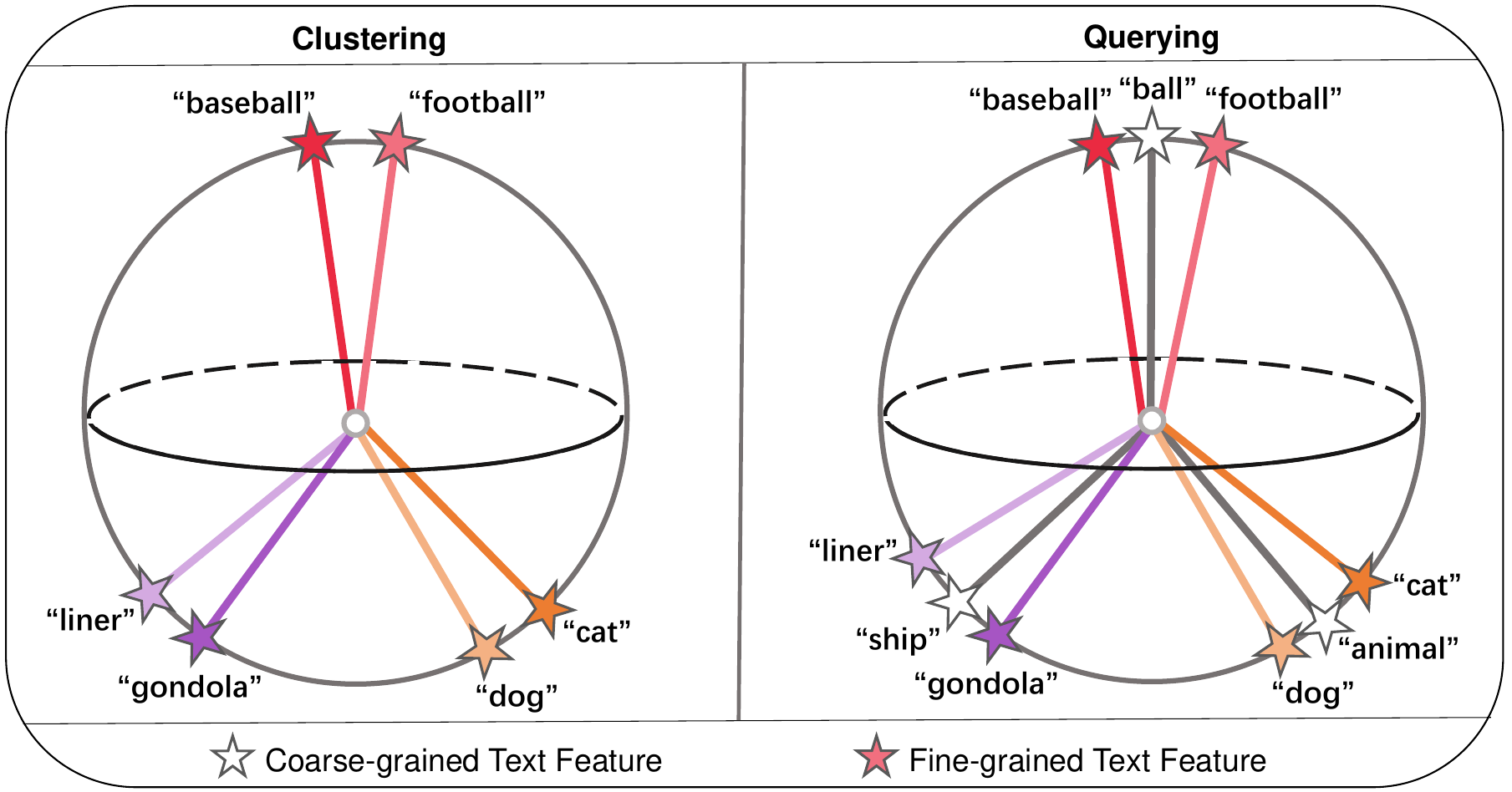}
		\caption{The overview of coarse semantic generation module (CSG). Here we take the categories of task as input data, and extract coarse-grained semantics. Applying KMeans++ \cite{ArthurV07} to text features of all categories to perform clustering (\textbf{Left}), then instruct LLM to perform semantic extraction on the texts within each cluster, extracting the common coarse-grained categories to which each cluster belongs (\textbf{Right}).}
		\label{fig:ex}
		\vspace{-10pt}
	\end{figure}
	In PromptStyler, the model necessitates the semantic consistency loss with all categories and demonstrates remarkable performance in less-category datasets. However, we observe that with the increase of categories, the performance tends to deteriorate. 
	In multi-category scenarios, many semantics have a high similarity, and redundant semantic information greatly compresses the learning space for style diversity, limiting the diversity visible to model.
		
	The objective is to enhance the style diversity , which we model it as an optimization problem, as follows:
	\begin{equation}
		\label{eq:1}
		\min_{\theta}(L(\theta) + \lambda R(\theta)), 
	\end{equation}
	where $L(\theta)$ is our objective function: style diversity loss, $\theta$ is parameter of learnable pseudo-style word embeddings, $\lambda$ is regularization coefficient, and $R(\theta)$ is regularization: semantic consistency loss. According to PromptStyler, we can define $L(\theta)$ and $R(\theta)$ as:
	
	\begin{small}
	\begin{equation}
		L(\theta) = \sum_{i=1}^{K} \sum_{j=1}^{i-1} |cos<f_{style}(\theta_i), f_{style}(\theta_j)>|, 
		\label{eq:2}
	\end{equation}
	\begin{equation}
		R(\theta) = \sum_{i=1}^{K} \sum_{c=1}^{N} -cos<f_{content}(c), f_{style-content}(\theta_i, c)>,
		\label{eq:3}
	\end{equation}	
	\end{small}
	\hspace{-0.3cm} where the \textit{K} is number of pseudo-styles and \textit{N} is the number of categories. Note that three kinds of prompts we used: a style prompt (``a \textit{\textbf{S$_i$}} style of a''), a content prompt (``[class]$_c$''), and a style-content prompt (``a \textit{\textbf{S$_i$}} style of a [class]$_c$''), and $\theta_i$ is parameter of i-th pseudo-style word embedding. $f_{style}$, $f_{content}$ and $f_{style-content}$ is style prompt feature, content prompt feature and style-content feature respectively encoded and normalized by CLIP's text encoder. The $cos<\cdot>$ stands for the cosine calculation operation. From Eq. \ref{eq:3}, it can be seen that the strength of semantic consistency constraint is proportional to the number of categories \textit{N}. 
	Thus, the core issue to enhance style diversity lies in reducing \textit{N}, but directly reducing \textit{N} would lead to a corresponding weakness of semantic consistency, causing styles to learn completely at random. Therefore, it is necessary to retain semantic consistency as much as possible while reducing \textit{N}.
	
	When the number of categories \textit{N} is very large, there will be a high similarity among some categories (\eg ~`tabby cat' and `tiger cat'), which implies the existence of redundant semantics. We cluster similar semantic information into piles and then extract coarse-grained semantics from each pile to eliminate redundant information, achieving the goal of reducing \textit{N}, thereby enhancing style diversity. Next, we will introduce proposed module \textbf{C}oarse \textbf{S}emantic \textbf{G}eneration \textbf{(CSG)} to address this issue.
	
	Based on above observation, we design an automatic extraction module to extract common semantics. Specifically, we encode categories by text encoder to derive the text features. Then, we apply KMeans++ \cite{ArthurV07} to cluster these text features, and the metric of clustering we used is silhouette coefficient, which can determine the optimal number of clusters without manual intervention. Text within each cluster can be regarded as sharing some coarse-grained semantics, as shown in Fig. \ref{fig:ex}. Next, we extract the common coarse-grained semantics from each cluster. 
	
	Inspired by GPT's powerful capabilities, we utilize GPT-4 \cite{achiam2023gpt} to extract coarse-grained semantics. Specifically, for a cluster, we query GPT-4 with: 
	\begin{center}
		\label{text:query}
		\texttt{"Q: Tell me $\left \{ list \ of \ category \ names \ under \ a \ cluster \right \}$ have in common with three words. If not, it can be nothing."}. 
	\end{center}
	After extraction, we further check if generated coarse-grained semantics contains "nothing", it means that no common information. If so, we perform binary clustering for corresponding cluster and query again. It should be noted that although we considered the scenario where no shared semantics exist, in actual execution, we have never encountered ``nothing". We define this \textbf{c}oarse \textbf{s}emantic \textbf{s}et as \textbf{\textit{css}}. Based on this process, when dealing with numerous categories, we can perform semantic consistency loss using few common semantic (\eg ~ [``tabby cat", ``tiger cat"] = ``cat"). We leverage coarse-grained consistency loss $\mathcal{L}_{SC}$ to replace semantic consistency, as formulated below:
	
	\begin{small}
		\begin{align}
			\mathcal{L}_{SC}=\sum_{i=1}^{K} \sum_{c \in \mathbf{css}} -cos<f_{content}(c), f_{style-content}(\theta_i, c)>,
		\end{align}
	\end{small}
	where \textit{c} is from \textbf{\textit{css}}, instead of original categories. Using few coarse-grained category names to guarantee semantic consistency, we effectively avoid disturbing the diversity of styles. With this naive extension, the performance has been significantly benefited in multi-category scenarios. 
	\vspace{-0.2cm}
	
	\subsection{Uniform Style Generation}
	\label{section: uniform generation}
	In \cite{Cho_2023_ICCV}, the style diversity is trained by orthogonality, which means there must be a large portion of joint space that remains uncovered if the number of pseudo-style \textit{K} is small than the dimension of joint space (\eg ~80 orthogonal vectors can not cover a 1024-dimension space according to schmidt orthogonalization). The core issue is that style diversity is learned merely based on orthogonality. Therefore, we propose a module: \textbf{U}niform \textbf{S}tyle \textbf{G}eneration \textbf{(USG)}, which uses a uniformly distributed template to learn more diverse styles. Inspired by the characteristic of neural collapse, it can initialize a set of vectors with equal and maximum margin, which we use to initialize the templates. We employ these vectors initialized by neural collapse as template to enhance the style diversity. In addition, we present this set of templates in the form of a classifier, which conveniently enables parallel training of styles, thereby improving training efficiency.
	
	Specifically, we pre-define \textit{K} vectors by neural collapse, where the \textit{K} is number of pseudo-styles, which is used as the initial template for pseudo-styles. Each pseudo-style has its own unique template vector, which is trained through semantic consistency and template. We assume that \textbf{\^{W}} = [$w_1$,...,$w_K$] $\in \mathbb{R}^{P \times K}$ as the vectors initialized by neural collapse, where \textit{P} is dimension of vision-language model joint space. Each pair within \textbf{\^{W}} satisfies Eq. \ref{equation:neural collapse}:
	\begin{align}
		w^T_{n1}w_{n2} = \frac{K}{K-1} \delta_{n1, n2} - \frac{1}{K-1}, \forall n_1,n_2 \in [1,K], 
		\label{equation:neural collapse}
	\end{align} 
	where $\delta_{n1,n2}$=1 when $n_1$ = $n_2$ and 0 otherwise. Therefore, they have equal and maximal margin between any pairwise vectors. Most importantly, they are uniformly distributed throughout the entire joint space.
	
	In order to improve the efficiency, we combine these vectors into a fixed classifier, which allows for parallel training. Simultaneously, to integrate the training of style diversity, we assign labels to each template in sequence from 1 to \textit{K}. With this fixed classifier and the labels we manually assigned, we can achieve diversity training through a simple cross-entropy loss $\mathcal{L}_{CE}$. Not only does it discard complex loss, but it also allows for more uniform diversity.
	\vspace{0.4cm}

	\subsection{Training and Inference}
	In training phase, the whole training process is divided into two stages. The first stage is the style generation, which is used to generate data for training of next stage. After the style generation, we take \textit{$f_{style-content}$} as training data and train a linear classifier \textit{F}$_C$ to obtain the final class prediction. Notably, this classifier differs from the fixed classifier in Sec. \ref{section: uniform generation}. The training of first stage can be trained with: 
	\vspace{-0.1cm}
	\begin{align}
		\label{formula:1}
		\mathcal{L}_{1\text{-}stage} = \mathcal{L}_{CE} + \mathcal{L}_{SC}.
	\end{align}
	Similar to \cite{Cho_2023_ICCV}, we employ ArcFace Loss \cite{DengGXZ19} as our classification loss $\mathcal{L}_C$ in second training stage. Crucially, we employ coarse-grained semantics to guarantee semantic consistency in first training stage, while in second stage, we utilize fine-grained semantics to train classifier. 
	
	In the inference stage, we simply use image encoder to replace text encoder. Given an image, the image encoder encode image to feature in joint vision-language space, and then used for classification. 
	
	\subsection{Discussion}
	\label{sec:dis}
	Our method is designed to enhance the diversity of synthesized styles in multi-category scenarios. In Sec. \ref{sec:csg}, we have formulated it as an optimization problem, as formulated in Eq. \ref{eq:1}. It can be seen that there are two approaches to weakening semantic consistency constraints: reducing \textit{N} and lowering $\lambda$. In proposed method, we adopt the strategy of reducing \textit{N}. In this section, we discuss why we do not directly lower $\lambda$. To evaluate, we make modification to the loss of \cite{Cho_2023_ICCV}, as follows:
	\begin{equation}
		\mathcal{L}_{prompt} = \mathcal{L}_{style} + \lambda \mathcal{L}_{content}, 
		\label{eq:7}
	\end{equation}
	where $\lambda$ is between 0 and 1. As seen in Tab. \ref{tab:lamda}, with the increase of $\lambda$, style diversity has indeed rise (the lower the SD, the better style diversity), but semantic consistency
	is also continuing to drop (the higher the SC, the better the semantic consistency), which means increasingly
	poor semantic consistency. Therefore, we conclude that it is not possible to directly enhance style diversity while
	maintaining semantic information by simply reducing the semantic consistency loss coefficient $\lambda$. This leads to
	the development of our method: by extracting coarse-grained semantics to reduce \textit{N}, we weaken the overly strong constraints of semantic consistency on style diversity, thereby expanding the space of style diversity.

	\section{Experiments}
	\label{sec:experiment}
	\renewcommand{\cmidrulesep}{0mm} 
	\setlength{\aboverulesep}{0mm} 
	\setlength{\belowrulesep}{0mm} 
	\setlength{\abovetopsep}{0cm}  
	\setlength{\belowbottomsep}{0cm}

	\newcommand{\PreserveBackslash}[1]{\let\temp=\\#1\let\\=\temp}
	\newcolumntype{C}[1]{>{\PreserveBackslash\centering}p{#1}}
	\newcolumntype{R}[1]{>{\PreserveBackslash\raggedleft}p{#1}}
	\newcolumntype{L}[1]{>{\PreserveBackslash\raggedright}p{#1}}
	
	\begin{table*}[htbp]
		\centering
		\caption{Comparison with state-of-the-art domain generalization methods, ZS-CLIP (C) refers to zero-shot CLIP using `[class]' as its text prompt, and ZS-CLIP (PC) indicates zero-shot CLIP using `a photo of a [class]' as its text prompt. 
			\textbf{Bold} indicates the best result, and the gray line is our method. M-Avg denotes the average result across multi-category datasets, while Avg is the average result across all datasets. \textcolor{red}{`\dag'} denotes that the method uses images as training data. Some methods do not provide standard errors, so we set stand errors of these methods to ±0.0 directly.}
		\renewcommand{\arraystretch}{1.75}
		\resizebox{\textwidth}{!}{
			\begin{tabular}{l|c|ccc|ccc|c} 
				\toprule
				\multirow{3}{2.5cm}{\Large Method} & \multirow{3}{2cm}{\Large \centering Venue}  & \multicolumn{7}{c}{\Large Accuracy (\%)$\uparrow$}\\
				\cline{3-9}
				& & \multicolumn{3}{c|}{\Large Less-category Benchmark} & \multicolumn{3}{c|}{\Large Multi-category Benchmark} & \multirow{2}{*}{\Large Avg./M-Avg.}\\
				\cline{3-8}
				& & \Large PACS & \Large VLCS & \Large OfficeHome & \Large ImageNet-R & \Large DomainNet & \Large ImageNet-S & \\ 
				\midrule
				\multicolumn{9}{c}{\Large ResNet-18 with pre-trained weights on ImageNet \rule{0pt}{0.5em}} \\ 
				\midrule
				\Large SODG-NET \cite{BeleBBJRB24}\textcolor[rgb]{1, 0, 0}{$^\dag$} & \Large WACV'2024 & \quad {\Large -} & \quad {\Large -} & \quad {\Large 61.4}{±0.0} & \Large - & {\Large -} & \Large - & \Large - \\
				\midrule
				\multicolumn{9}{c}{\Large ResNet-50 with pre-trained weights on ImageNet \rule{0pt}{1em}} \\ 
				\midrule
				\Large MSAM \cite{LiLWLT23}\textcolor{red}{$^\dag$} & \Large TCSVT'2023 & \quad {\Large 88.5}{±0.0} & \quad {\Large 77.3}{±0.0} & \quad {\Large 65.6}{±0.0} & \Large - & {\Large 48.0}{±0.0} & \Large - & \Large - \\
				\Large NormAUG \cite{QiYSG24}\textcolor{red}{$^\dag$} & \Large TIP'2024 & \quad {\Large 89.5}{±0.0} & \quad \Large - & \quad {\Large 67.8}{±0.0} & \Large - & {\Large 46.5}{±0.0} & \Large - & \Large - \\
				\Large IPCL \cite{ChenWZSMD24}\textcolor{red}{$^\dag$} & \Large TCSVT'2024 & \quad {\Large 86.6}{±0.0} & \quad {\Large 78.3}{±0.0} & \quad {\Large 67.7}{±0.0} & \Large - & \Large - & \Large - & \Large - \\
				\Large SFADA \cite{HeWTL24} & \Large PR'2024 & \quad {\Large 94.0}{±0.0} & \quad {\Large -} & \quad {\Large 71.9}{±0.0} & \Large - & {\Large 51.9}{±0.0} & \Large - & \Large - \\
				\Large NAMI \cite{ZhangWSZLZ24}\textcolor{red}{$^\dag$} & \Large TMM'2024 & \quad {\Large -}- & \quad {\Large -} & \quad {\Large 75.8}{±0.0} & \Large - & \Large - & \Large - & \Large - \\
				\midrule
				\multicolumn{9}{c}{\Large ResNet-50 with pre-trained weights from CLIP \rule{0pt}{1em}} \\ 
				\midrule
				\Large ZS-CLIP(C)\cite{radford2021learning} & \Large ICML'2021 & \quad {\Large 91.3}{±0.0} & \quad {\Large 79.1}{±0.0} & \quad {\Large 70.6}{±0.0} & {\Large 58.8}{±0.0} & {\Large 45.9}{±0.0} & {\Large 32.4}{±0.0} & {\Large 63.0/45.7} \\
				\Large ZS-CLIP(PC)\cite{radford2021learning} & \Large ICML'2021 & \quad {\Large 91.1}{±0.0} & \quad {\Large 82.2}{±0.0} & \quad {\Large 71.5}{±0.0} & {\Large 56.3}{±0.0} & {\Large 46.1}{±0.0} & {\Large 32.1}{±0.0} &  {\Large 63.2/44.8}\\ 
				\Large WaffleCLIP\cite{RothKKVSA23} &  \Large ICCV'2023 & \quad{\Large 92.5}{±0.2} & \quad {\Large 82.7}{±0.5} & \quad {\Large 72.1}{±0.0} & {\Large 59.2}{±0.3} & {\Large 47.1}{±0.4} & {\Large 32.1}{±1.1} & {\Large 64.3/46.1} \\
				\Large PromptStyler\cite{Cho_2023_ICCV} & \Large ICCV'2023 & \quad {\Large 93.2}{±0.4} & \quad {\Large 83.2}{±0.2} & \quad {\Large 71.9}{±0.2} & {\Large 54.8}{±0.1} & {\Large 46.9}{±0.2} & {\Large 31.7}{±0.2} & {\Large 63.6/44.5} \\ 
				\Large STYLIP \cite{BoseJFS0B24}\textcolor[rgb]{1, 0, 0}{$^\dag$} & \Large WACV'2024 & \quad {\Large 92.6}{±0.0} & \quad {\Large 77.2}{±0.2} & \quad {\Large 71.6}{±0.0} & \Large - & \Large - & \Large - & \Large - \\ 
				\Large PromptTA \cite{zhang2024promptta} & \Large - & \quad \textbf{{\Large 93.8}{±0.0}} & \quad {\Large 83.2}{±0.1} & \quad \textbf{{\Large 73.2}{±0.1}} & {\Large 58.3}{±0.4} & {\Large 46.9}{±0.2} & {\Large 32.3}{±1.7} & \Large 64.6/45.8 \\
				\Large DPStyler \cite{tang2024dpstyler} & \Large TMM'2024 & \quad {\Large 93.6}{±0.2} & \quad \textbf{{\Large 83.5}{±0.2}} & \quad {\Large 72.5}{±0.2} & {\Large 57.4}{±1.7} & \textbf{{\Large 47.9}{±0.1}} & {\Large 32.2}{±0.4} & \Large 64.5/45.9 \\ 
				\rowcolor{gray!20} \Large BatStyler & \Large Ours & \quad {\Large 93.2}{±0.2}   &  \quad {\Large 83.2}{±0.2}  &  \quad {\Large 72.4}{±0.4} & \textbf{{\Large 59.9}{±0.2}}  &  {\Large 47.8}{±0.1} & \textbf{{\Large 32.9}{±0.1}}  & \makebox[0.1\textwidth][c]{\textbf{{\Large 64.9/46.9}}}   \\ 
				\midrule
				\multicolumn{9}{c}{\Large ViT-B/16 with pre-trained weights from CLIP \rule{0pt}{1em}} \\
				\midrule
				\Large ZS-CLIP(C)\cite{radford2021learning} & \Large ICML'2021 & \quad {\Large 95.7}{±0.0} & \quad {\Large 81.9}{±0.0} & \quad {\Large 80.8}{±0.0} & {\Large 75.1}{±0.0} & {\Large 57.4}{±0.0} & {\Large 44.9}{±0.0} & {\Large 72.6/59.1} \\
				\Large ZS-CLIP(PC)\cite{radford2021learning} & \Large ICML'2021 & \quad {\Large 95.8}{±0.0} &  \quad {\Large 82.2}{±0.0} & \quad {\Large 81.8}{±0.0} & {\Large 73.6}{±0.0} & {\Large 57.2}{±0.0} & {\Large 44.9}{±0.0} & {\Large 72.6/58.6} \\
				\Large WaffleCLIP\cite{RothKKVSA23} & \Large ICCV'2023 & \quad {\Large 96.7}{±0.1} & \quad {\Large 82.4}{±0.5} & \quad {\Large 82.4}{±0.5} & {\Large 76.3}{±0.2} & {\Large 57.8}{±0.1} & {\Large 45.3}{±0.4} & {\Large 73.5/59.8} \\
				\Large PromptStyler\cite{Cho_2023_ICCV} & \Large ICCV'2023 & \quad \textbf{{\Large 97.5}{±0.2}} & \quad {\Large 82.9}{±0.5} & \quad {\Large 82.9}{±0.5} & {\Large 71.9}{±0.4} & {\Large 56.7}{±0.3} & {\Large 43.9}{±0.3} & {\Large 72.6/57.5} \\ 
				\Large STYLIP \cite{BoseJFS0B24}\textcolor[rgb]{1, 0, 0}{$^\dag$} & \Large WACV'2024 & \quad {\Large 97.0}{±0.0} & \quad {\Large 82.9}{±0.0} & \quad \textbf{{\Large 83.9}{±0.0}} & \Large - & \Large - & \Large - & \Large - \\ 
				\Large DCLIP \cite{MenonV23} & \Large ICLR'2023 & \quad {\Large 91.4}{±0.4} & \quad {\Large 81.3}{±0.2} & \quad {\Large 81.9}{±0.1} & {\Large 73.0}{±0.1} & {\Large 56.6}{±0.5} & {\Large 45.7}{±0.2} & {\Large 71.6/58.4} \\
				\Large Cp-CLIP \cite{RenS023} & \Large NeurIPS'2023 & \quad {\Large 96.5}{±0.1} & \quad {\Large 82.7}{±0.4} & \quad {\Large 79.9}{±0.0} & \textbf{{\Large 76.9}{±0.1}} & {\Large 57.2}{±0.4} & {\Large 43.2}{±2.1} & {\Large 72.7/59.1} \\
				\Large PromptTA \cite{zhang2024promptta} & \Large - & \quad {\Large 97.3}{±0.1} & \quad {\Large 83.6}{±0.3} & \quad {\Large 82.9}{±0.1} & {\Large 75.8}{±0.2} & {\Large 57.2}{±0.7} & {\Large 44.3}{±0.2} & \Large 73.5/59.9 \\
				\Large DPStyler \cite{tang2024dpstyler} & \Large TMM'2024 & \quad {\Large 97.1}{±0.1} & \quad \textbf{{\Large 84.0}{±0.4}} & \quad {\Large 82.8}{±0.1} & {\Large 76.1}{±0.2} & \textbf{{\Large 58.9}{±0.1}} & {\Large 45.0}{±0.3} & \Large 73.9/59.8 \\ 
				\rowcolor{gray!20} \Large BatStyler & \Large Ours & \quad {\Large 97.3}{±0.1} & \quad {\Large 82.7}{±0.4} & \quad {\Large 83.7}{±0.2} & {\Large 76.6}{±0.1} & {\Large 58.5}{±0.4} & \textbf{{\Large 45.9}{±0.1}} &  \makebox[0.1\textwidth][c]{\textbf{{\Large 74.1/60.3}}}   \\ 
				\midrule
				\multicolumn{9}{c}{\Large ViT-L/14 with pre-trained weights from CLIP \rule{0pt}{1em}} \\ 
				\midrule
				\Large ZS-CLIP(C)\cite{radford2021learning} & \Large ICML'2021 &\quad  {\Large 97.9}{±0.0} & \quad {\Large 78.3}{±0.0} & \quad {\Large 86.6}{±0.0} & {\Large 85.9}{±0.0} & {\Large 63.1}{±0.0} & {\Large 55.8}{±0.0} & {\Large 77.9/68.3} \\
				\Large ZS-CLIP(PC)\cite{radford2021learning} &  \Large ICML'2021 & \quad {\Large 98.4}{±0.0} & \quad {\Large 81.9}{±0.0} & \quad {\Large 86.2}{±0.0} & {\Large 85.3}{±0.0} & {\Large 63.0}{±0.0} & {\Large 56.4}{±0.0} &  {\Large 78.5/68.2} \\
				\Large WaffleCLIP\cite{RothKKVSA23} &  \Large ICCV'2023 & \quad {\Large 98.5}{±0.2} & \quad {\Large 82.2}{±0.2} & \quad {\Large 87.1}{±0.2} & {\Large 86.1}{±0.1} & {\Large 63.7}{±0.8} & {\Large 56.4}{±0.1} & {\Large 79.0/68.7} \\
				\Large PromptStyler\cite{Cho_2023_ICCV} &  \Large ICCV'2023 & \quad \textbf{{\Large 98.6}{±0.1}}  & \quad \textbf{{\Large 82.4}{±0.3}} & \quad \textbf{{\Large 87.9}{±0.3}} & {\Large 82.3}{±0.2} & {\Large 62.9}{±0.6} & {\Large 55.1}{±0.2} & {\Large 78.2/66.8} \\ 
				\rowcolor{gray!20} \Large BatStyler & \Large Ours & \quad {\Large 98.4}{±0.1} & \quad {\Large 82.2}{±0.1} & \quad {\Large 87.4}{±0.04} & \textbf{{\Large 86.8}{±0.3}} & \textbf{{\Large 64.4}{±0.3}} & \textbf{{\Large 56.8}{±0.1}} &  \makebox[0.1\textwidth][c]{\textbf{{\Large 79.3/69.3}}}              \\
				\bottomrule
			\end{tabular}	
		}
		\label{table:main}
		\vspace{-0.6em}
	\end{table*}

	In this section, we begin by presenting the datasets in Section~\ref{sec:EXP-DS} and implementation details in Sec. \ref{sec:EXP-ID}. Following that, we evaluate our proposed method against the current state-of-the-art domain generalization source-free image classification techniques in Section~\ref{sec:EXP-CUA}. To verify the impact of different modules in our framework, we carry out ablation studies in Section~\ref{sec:EXP-SS}. Finally, we delve deeper into the properties of our approach in Section~\ref{sec:EXP-FA}.
	\vspace{-0.1cm}
	\subsection{Evaluation Datasets}
	\label{sec:EXP-DS} 
	
	To evaluate the effectiveness of our method, we conduct experiments on six domain generalization benchmark datasets: PACS \cite{LiYSH17},  VLCS \cite{FangXR13}, OfficeHome \cite{VenkateswaraECP17}, ImageNet-R \cite{HendrycksBMKWDD21}, DomainNet \cite{PengBXHSW19} and ImageNet-S \cite{GaoLYCHT23}. The \textbf{PACS} dataset consists of $7$ classes captured from four domains: art, cartoon, photo and sketch, and the validation set has $9991$ images. The \textbf{VLCS} dataset consists of $5$ classes from four domains: VOC2007, LabelMe, SUN09 and Caltech101, totaling approximately 25K images. The \textbf{OfficeHome} dataset has $65$ classes from four domains: art, clipart, product and real-world, accumulating to a total of 15,500 images. The \textbf{ImageNet-R} dataset has $200$ classes from many domains (\eg ~sculptures, art), which has about 30K images. The \textbf{DomainNet} datasets has 345 classes from six different domains, which has about 600K images. Finally, the \textbf{ImageNet-S} dataset has $1000$ classes from many domains in ImageNet, which has about 1.2 million images for evaluation. According to the task definition of source-free domain generalization, we do not use any images for training, hence, all images are reserved for evaluation. Note that we define PACS, VLCS and OfficeHome as less-category datasets, and ImageNet-R, DomainNet and ImageNet-S as multi-category datasets with more categories.
	
	\subsection{Implementation Details} 
	\label{sec:EXP-ID}
	BatStyler maintains consistent implementation and training with identical configurations across all evaluation datasets. The models are trained on a single RTX3090 GPU. Further detailed comparisons are detailed as follows: 
	
	\textbf{Architecture.} We employ CLIP \cite{radford2021learning} as vision-language model, and implement with ResNet-50, ViT-B/16 and ViT-L/14 for evaluation. Notably, the image encoder and text encoder are all frozen during training. The dimension of their joint space 1024, 512, 768 respectively, and the dimension of embedding are 512, 512, 768 respectively.
	
	\textbf{Coarse Semantic Generation.} Before training, we employ silhouette coefficient as clustering metric, the number of clusters is set to the value where the silhouette coefficient reaches its peak value. Additionally, we employ GPT-4 to extract 3 coarse-grained semantics for each cluster. It should be noted that this module is function before training.
	
	\textbf{Style Generation.} We integrate vectors initialized by neural collapse as a fixed classifier, and generate pseudo-styles in parallel. We train the styles for 300 epochs using SGD optimizer with a learning rate 0.2, a momentum 0.9, a batch size 4 and cosine scheduler to adjust dynamic learning rate. Note that the number of pseudo-styles \textit{K} must be divisible by batch size. The default number of pseudo-style word vectors is \textit{K}=80. Only the style embedding are trainable in this stage.
	\vspace{-0.3cm}
	
	\textbf{Linear classifier training.} We initialize linear classifier with normal distribution and train the classifier for 50 epochs using SGD optimizer 0.01, a momentum 0.9 and a batch size 128. Similarly, we utilize ArcFace \cite{DengGXZ19} as loss with scaling factor 5 and angular margin 0.5. The training data are well-trained style-content prompt in style generation.
	
	\textbf{Inference.} Given an image, we resize it to 224$\times$224 and normalize. In the end, we select the max score as prediction.
	\vspace{-0.5cm}
	\begin{table*}[t]
		\centering
		\caption{Accuracy on each domain. The experiments are implemented on DomainNet with ResNet-50. The domains are art painting, cartoon, photo and sketch. \textbf{Bold} indicates the best result, and the gray line is our method.}
		\renewcommand{\arraystretch}{1.0}
		\resizebox{\textwidth}{!}{
			\begin{tabular}{l|c|cccccc|c}
				\toprule
				\multirow{2}{2cm}{Method} & \multirow{2}{2cm}{\centering Venue} & \multicolumn{7}{c}{Accuracy (\%)} \\
				\cmidrule{3-9} & & Clipart & Infograph & Painting & Quickdraw & Real & Sketch & Avg. \\
				\midrule
				\multicolumn{9}{c}{ResNet-50 with pre-trained weights from CLIP} \\
				\midrule
				ZS-CLIP (C) \cite{radford2021learning} & ICML'2021 & 52.8\tiny±0.0 & 40.1\tiny±0.0 & 52.9\tiny±0.0 & 6.5\tiny±0.0 & 75.3\tiny±0.0 & 47.6\tiny±0.0 & 45.9 \\
				ZS-CLIP (PC) \cite{radford2021learning} & ICML'2021 & 53.1\tiny±0.0 & 39.4\tiny±0.0 & 52.9\tiny±0.0 & 5.7\tiny±0.0 & 76.6\tiny±0.0 & 48.5\tiny±0.0 & 46.1 \\
				WaffleCLIP \cite{RothKKVSA23} & ICCV'2023 & 54.0\tiny±0.1 & 40.3\tiny±0.7 & 54.2\tiny±0.4 & 6.6\tiny±0.9 & \textbf{77.9}\tiny±0.1 & 49.3\tiny±0.0 & 47.0 \\
				PromptStyler \cite{Cho_2023_ICCV} & ICCV'2023 & 53.9\tiny±0.2 & \textbf{41.4}\tiny±1.1 & 54.6\tiny±0.2 & 5.6\tiny±0.7 & 76.8\tiny±0.1 & 49.2\tiny±0.2 & 46.9 \\
				\rowcolor{gray!20} BatStyler & Ours & \textbf{55.2}\tiny±0.4 & 41.3\tiny±0.1 & \textbf{55.8}\tiny±0.1 & \textbf{6.8}\tiny±0.2 & 77.3\tiny±0.1 & \textbf{50.4}\tiny±0.4 & \textbf{47.8} \\
				\midrule
				\multicolumn{9}{c}{ViT-B/16 with pre-trained weights from CLIP} \\
				\midrule
				ZS-CLIP (C) \cite{radford2021learning} & ICML'2021 & 70.2\tiny±0.0 & 48.9\tiny±0.0 & 65.7\tiny±0.0 & 14.3\tiny±0.0 & 82.4\tiny±0.0 & 62.7\tiny±0.0 & 57.4 \\
				ZS-CLIP (PC) \cite{radford2021learning} & ICML'2021 & 70.4\tiny±0.0 & 47.3\tiny±0.0 & 65.0\tiny±0.0 & 13.5\tiny±0.0 & 83.3\tiny±0.0 & 63.6\tiny±0.0 & 57.2 \\
				WaffleCLIP \cite{RothKKVSA23} & ICCV'2023 & 70.9\tiny±0.1 & 49.7\tiny±0.5 & 66.3\tiny±0.0 & 15.4\tiny±0.1 & 82.7\tiny±0.1 & 63.4\tiny±0.0 & 58.1 \\
				PromptStyler \cite{Cho_2023_ICCV} & ICCV'2023 & 70.1\tiny±0.5 & 47.4\tiny±0.3 & 65.1\tiny±0.3 & 12.5\tiny±0.7 & 82.3\tiny±0.1 & 62.4\tiny±0.9 & 56.7 \\
				\rowcolor{gray!20} BatStyler & Ours & \textbf{71.1}\tiny±0.6 & \textbf{50.2}\tiny±0.1 & \textbf{66.4}\tiny±0.3 & \textbf{15.8}\tiny±0.2 & \textbf{83.4}\tiny±0.1 & \textbf{64.0}\tiny±0.4 & \textbf{58.5} \\
				\midrule
				\multicolumn{9}{c}{ViT-L/14 with pre-trained weights from CLIP} \\
				\midrule
				ZS-CLIP (C) \cite{radford2021learning} & ICML'2021 & 77.6\tiny±0.0 & 52.7\tiny±0.0 & 71.0\tiny±0.0 & 21.6\tiny±0.0 & 85.9\tiny±0.0 & 70.0\tiny±0.0 & 63.1 \\
				ZS-CLIP (PC) \cite{radford2021learning} & ICML'2021 & 78.3\tiny±0.0 & 50.6\tiny±0.0 & 69.0\tiny±0.0 & 22.4\tiny±0.0 & 86.3\tiny±0.0 & \textbf{71.5}\tiny±0.0 & 63.0 \\
				WaffleCLIP \cite{RothKKVSA23} & ICCV'2023 & \textbf{78.4}\tiny±0.1 & 53.7\tiny±0.5 & 71.8\tiny±0.0 & 22.4\tiny±0.1 & \textbf{86.7}\tiny±0.1 & 70.9\tiny±0.0 & 63.9 \\
				PromptStyler \cite{Cho_2023_ICCV} & ICCV'2023 & 77.5\tiny±0.5 & 52.3\tiny±0.7 & 70.8\tiny±1.4 & 21.0\tiny±0.7 & 86.1\tiny±0.1 & 69.5\tiny±0.4 & 62.9 \\
				\rowcolor{gray!20} BatStyler & Ours & 78.0\tiny±0.1 & \textbf{54.2}\tiny±0.4 & \textbf{73.2}\tiny±0.3 & \textbf{23.4}\tiny±0.2 & 86.5\tiny±0.1 & \textbf{71.5}\tiny±0.4 & \textbf{64.4} \\
				\bottomrule
			\end{tabular}
			\label{table:domainnet}
			\vspace{-10pt}
		}
	\end{table*}
	
	\begin{table*}[t]
	\centering
	\small
	\caption{The evaluation of training and inference resources on GPU memory usage, training time, model parameter count and inference speed. Training Stage 1 represents the style generation process, and Training Stage 2 denotes the training of classifier. `USG' stands for Uniform Style Generation, while `w/' and `w/o' respectively indicate whether USG is used. }
	\renewcommand{\arraystretch}{1.0}
	\begin{tabular}{l|cc|cc|cc} 
		\toprule
		\multirow{2}{*}{Method} & \multicolumn{2}{c|}{Training Stage 1} & \multicolumn{2}{c|}{Training Stage 2} & \multicolumn{2}{c}{Inference}\\
		\cline{2-7}
		& Memory $\downarrow$ & Time $\downarrow$ & Memory $\downarrow$ & Time $\downarrow$ & Params $\downarrow$ & FPS $\uparrow$ \\
		\hline
		\multicolumn{7}{c}{ImageNet-R (200 classes) \rule{0pt}{0.5em}} \\
		\hline
		ZS-CLIP \cite{radford2021learning} & 0GB & 0s & 0GB & 0s & 102.01M & 72.6 \\
		PromptStyler \cite{Cho_2023_ICCV} & 5.57GB & 21.2min & 2.25GB & 13.8min & 36.73M & 90.4 \\
		\hline
		BatStyler w/o USG &  4.27GB & 14.6min & 1.90GB & 11.5min & 36.73M & 94.9 \\
		BatStyler w/ USG & 7.91GB & 2.4min & 1.48GB & 20.3min & 36.73M & 92.3 \\
		\hline
		\multicolumn{7}{c}{DomainNet (345 classes) \rule{0pt}{0.5em}} \\
		\hline
		ZS-CLIP \cite{radford2021learning} & 0GB & 0s & 0GB & 0s & 102.01M & 77.1 \\
		PromptStyler \cite{Cho_2023_ICCV} & 8.31GB & 29.0min & 2.16GB & 29.7min & 37.3M & 81.1 \\
		\hline
		BatStyler w/o USG & 3.80GB & 18.2min & 2.59GB & 33.8min & 37.3M & 85.3 \\
		BatStyler w/ USG & 7.29GB & 14.9min & 2.44GB & 36.4min & 37.3M & 82.9 \\
		\hline
		\multicolumn{7}{c}{ImageNet-S (1000 classes) \rule{0pt}{0.5em}} \\
		\hline
		ZS-CLIP \cite{radford2021learning} & 0GB & 0s & 0GB & 0s & 102.01M & 48.4 \\
		PromptStyler \cite{Cho_2023_ICCV} & 21.26GB & 72.7min & 2.27GB & 322min & 39.86M & 79.6 \\
		\hline
		BatStyler w/o USG &  4.91GB & 39.1min & 2.48GB & 285min & 39.86M & 81.9 \\
		BatStyler w/ USG & 9.24GB & 7.4min & 2.42GB & 340min & 39.86M & 89.7 \\
		\hline
	\end{tabular}
	\label{tab:td}
	\end{table*}
		\begin{table*}[t]
	\centering
	\caption{Ablation study on the Uniform Style Generation module (USG) and Coarse Semantic Generation(CSG) across PACS, VLCS, OfficeHome, ImageNet-R, DomainNet and ImageNet-S on ResNet-50. \textbf{Bold} indicates the best result, and the gray line is our method.}
	\hspace{-0.37cm}
	\resizebox{0.99\textwidth}{!}{
		\renewcommand{\arraystretch}{1.0}
		\begin{tabular}{cc|cccccc|c}
			\toprule
			\multicolumn{2}{p{1.6cm}|}{\centering Module} & \multicolumn{7}{c}{Accuracy (\%)} \\
			\midrule
			~USG & ~CSG
			& ~PACS & ~VLCS & ~OfficeHome & ~ImageNet-R & ~DomainNet & ~ImageNet-S & Avg.\\
			\midrule
			~\textbf{-} & ~\textbf{-} & \enspace 93.2 & ~\textbf{83.2} & 71.9 & 54.8 & 46.9 & 31.7 & 63.6 \\
			~\ding{51} & ~\textbf{-} & \enspace \textbf{93.4} & ~83.1 & 71.2 & 55.1 & 47.4 & 31.7 & 63.7 \\
			~\textbf{-} & ~\ding{51} & \enspace 93.2 & ~\textbf{83.2} & 71.9 & 59.2 & 47.7 & 32.7 & 64.5 \\
			\rowcolor{gray!20} ~\ding{51} & ~\ding{51} & \enspace 93.2 & ~\textbf{83.2} & \textbf{72.4} & \textbf{59.9} & \textbf{47.8} & \textbf{32.9} & \textbf{64.9} \\
			\bottomrule
	\end{tabular}}
	\label{table:ablation}
	\vspace{-10pt}
\end{table*}
	\vspace{0.1cm}
	\subsection{Evaluations}
	\label{sec:EXP-CUA}
	
	\textbf{Comparison with state-of-the-art methods.} We compare our method with image-free state-of-the-art (SOTA) methods, including Zero-shot CLIP \cite{radford2021learning}, WaffleCLIP \cite{RothKKVSA23}, PromptStyler \cite{Cho_2023_ICCV}, PromptTA \cite{zhang2024promptta}, DPStyler \cite{tang2024dpstyler}, Cp-CLIP \cite{RenS023}, DCLIP \cite{MenonV23} and SFADA \cite{HeWTL24}. In addition, we also compare our method with other methods, noting that these methods require training with source domain images, including MSAM \cite{LiLWLT23}, NormAUG \cite{QiYSG24}, IPCL \cite{ChenWZSMD24} and NAMI \cite{ZhangWSZLZ24}. The CLIP utilize web scale image-text pairs to train a vision-language model with contrastive learning. We only evaluate zero-shot performance on six datasets above with two prompts: C (\eg ~``dog") and PC (\eg ~``a photo of dog"). WaffleCLIP observe that the average of many prompts is the main driver of enhanced performance, instead of fine-grained semantics. Therefore, it proposed to initialize many different prompts with random words, which is a non-training method. The PromptStyler proposed to employ textual prompt to represent corresponding image feature, and train a linear classifier with text feature, which is a source-free method and do not need any images for training. The PromptTA and DPStyler are improvements over PromptStyler, both are source-free domain generalization methods. For zero-shot CLIP, we only evaluate performance on benchmark datasets. For WaffleCLIP, we use the same backbone with official configuration to perform. In PromptStyler, PromptTA and DPStyler, we employ same configuration with BatStyler to conduct the comparison. Compared to methods that are trained with source domain images, under the same model architecture, the performance improvement of our method is also quite significant. Results can be seen in Tab. \ref{table:main}. The Avg is average result across all datasets, and M-Avg is average result across multi-category datasets. Remakably, BatStyler outperforms other methods across multi-category datasets, and achieves comparable performance on less-category datasets.  Besides, although our method is slightly poorer than other methods on PACS, VLCS and OfficeHome (less-category datasets), the average result of our method is best in all downstream tasks.
	
	\textbf{Domain-independent analysis.} In Tab. \ref{table:domainnet}, we conduct an independent analysis of the performance on each domain. The dataset used is DomainNet, and all are conducted on ResNet-50. It is evident that the performance of CLIP-based models is highly related to CLIP itself. The performance of CLIP on the Quickdraw domain is relatively poor, and other CLIP-based models also perform relatively poorly in this domain. It can also be seen that the performance of BatStyler is better than PromptStyler, especially in Quickdraw and Sketch.
	
		\begin{figure}[t]
		\centering
		\includegraphics[width=9cm]{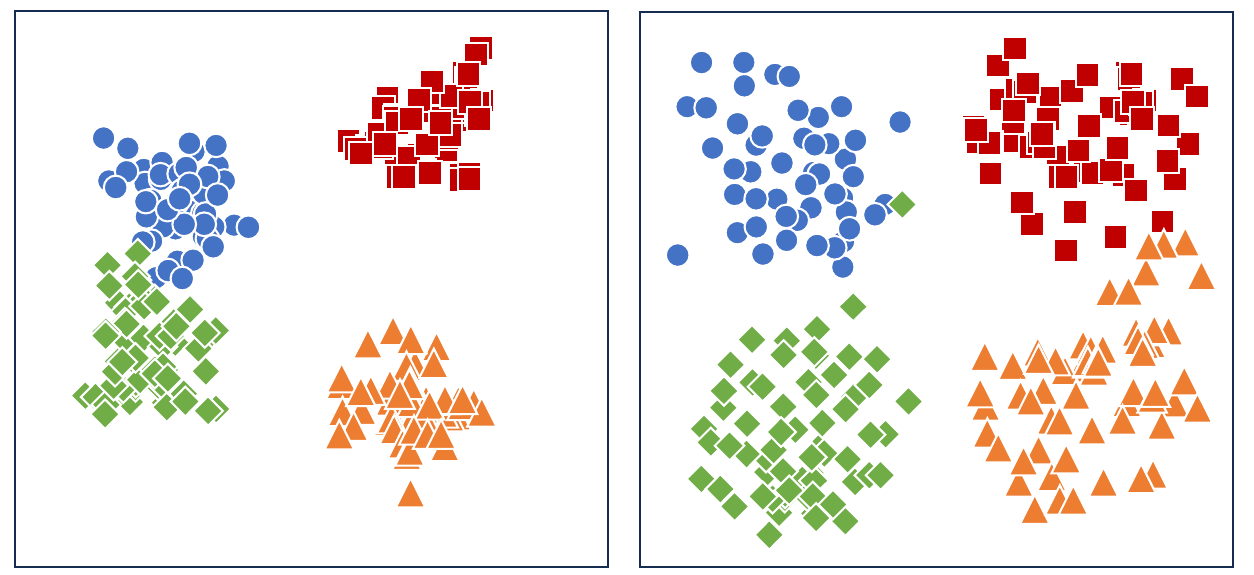}
		\caption{t-SNE \cite{van2008visualizing} visualization result. The style-content features visualized by PromptStyler (\textbf{Left}) and BatStyler (\textbf{Right}) on four randomly selected categories from ImageNet-S. Different colors represent different categories.} 
		\label{fig:tsne}
	\end{figure}

	\textbf{Evaluation on computation resources.} As shown in Tab. \ref{tab:td}, we provide detailed resource usage comparison between zero-shot CLIP \cite{radford2021learning}, PromptStyler \cite{Cho_2023_ICCV} and BatStyler (Ours) on multi-category datasets, encompassing both training and inference resources. From the table, it is noticeable that during training stage with USG, BatStyler's training time can be reduced to 10\% of PromptStyler in style generation. Due to the parallel training, the GPU memory is higher than other methods. It should be noted that we primarily focus on training stage 1: style generation.
	
		\begin{table}[t]
		\centering
		\caption{Comparison of accuracy under different classifier initialization strategy, \textbf{Random} denotes classifier with random initialization and not learnable, \textbf{Learnable} denotes that classifier can be trained, and \textbf{NC} denotes that classifier is initialized by neural collapse phenomenon and not learnable. Experiments are runs on ImageNet-S. \textbf{Bold} indicates the best result, and the gray line is our method.}
			\begin{tabular}{l|c|c|c}
				\toprule
				\multirow{2}{1.2cm}{Strategy} & \multicolumn{3}{c}{Accuracy(\%)} \\
				\cmidrule{2-4} & ResNet-50 & ViT-B/16 & ViT-L/14 \\
				\midrule
				Random & 31.2 & 44.7 & 53.2 \\
				Learnable & 32.4 & 45.7 & 56.1 \\
				\rowcolor{gray!20} NC & \textbf{32.9} & \textbf{45.9} & \textbf{56.8} \\
				\bottomrule
			\end{tabular}
		\label{table:classifier}%
	\end{table}%
			\begin{table}
		\centering
		\small
		\caption{Performance on different number of extracted semantics of each cluster \textit{C}. We run an evaluation with RN50 model of CLIP across multi-category benchmark datasets: ImageNet-R, DomainNet and ImageNet-S. \textbf{Bold} is the best result. }
		\setlength{\tabcolsep}{7pt}
		\renewcommand{\arraystretch}{1.2}
		\begin{tabular}{lcccc} 
			\toprule
			\textit{C} & 1 & 3 & 5 & 10 \\
			\hline
			ImageNet-R & 59.7 & 59.9 & \textbf{60.0} & 53.9 \\
			\hline
			DomainNet & 46.1 & \textbf{47.8} & 47.4 & 46.6 \\
			\hline
			ImageNet-S & 31.1 & 32.9 & \textbf{33.1} & 32.8 \\
			\bottomrule
		\end{tabular}
	\label{tab:N}
\end{table}
	
	\textbf{Evaluations of t-SNE visualization results.} We utilize style-content prompt to visualize style-content features, the different colors represent different categories.  Due to the vast number of categories of ImageNet-S, it is impractical to visualize all categories. Therefore, we randomly select four categories in ImageNet-S for visualization. The four randomly selected categories are mower, megalith, wing and airedale. From Fig. \ref{fig:tsne}, in multi-category scenarios, BatStyler exhibits better style diversity within a category, and preserves the intervals between different semantics.. 
	
	\subsection{Ablation Study}
	\label{sec:EXP-SS}
	
	\textbf{Main Ablation on modules.} We conduct ablation experiments to evaluate the effectiveness of each module in our proposed method. The results are presented in Tab. \ref{table:ablation}, where USG represents Uniform Style Generation and CSG represents Coarse Semantic Generation, respectively. From the table, when using USG or CSG individually, there is a significant performance improvement on multi-category datasets. The greater improvement with CSG suggests that the overly strong constraints imposed by semantic consistency are the main reason hindering the model's generalization to multi-category scenarios. Finally, the forth line, ``USG + CSG” module achieves the best average results in all tasks except PACS, confirming the effectiveness of the BatStyler.
		\begin{table*}[t]
		\centering
		\caption{Cosine similarity comparison on ImageNet-R. SD and SC refer to Style Diversity and Semantic Consistency respectively. The lower SD, the better style diversity; the higher SC, the better semantic consistency. $\lambda$ represents the loss coefficient of semantic consistency loss. }
		\renewcommand{\arraystretch}{1.2}
		\resizebox{\textwidth}{!}{
			\begin{tabular}{l|c|ccc|ccc|ccc|ccc}
				\toprule
				\multirow{3}{*}{Method} & \multirow{3}{*}{Metric} & \multicolumn{12}{c}{Cosine Similarity      } \\
				\cmidrule{3-14} & & \multicolumn{3}{c|}{ResNet-50} & \multicolumn{3}{c|}{ViT-B/32} & \multicolumn{3}{c|}{ViT-B/16} & \multicolumn{3}{c}{ViT-L/14} \\
				\cmidrule{3-14} & & $\lambda$=0.1 & $\lambda$=0.5 & $\lambda$=1.0 & $\lambda$=0.1 & $\lambda$=0.5 & $\lambda$=1.0 & $\lambda$=0.1 & $\lambda$=0.5 & $\lambda$=1.0 & $\lambda$=0.1 & $\lambda$=0.5 & $\lambda$=1.0 \\
				\midrule 
				\multirow{2}{*}{PromptStyler \cite{Cho_2023_ICCV}} & SD ($\downarrow$) & 0.12 & 0.27 & 0.64 & 0.13 & 0.41 & 0.71 & 0.12 & 0.21 & 0.82 & 0.06 & 0.19 & 0.77 \\
				& SC ($\uparrow$) & 0.17 & 0.36 & 0.42 & 0.21 & 0.34 & 0.35 & 0.3 & 0.37 & 0.39 & 0.11 & 0.21 & 0.38 \\
				\hline
				\multirow{2}{*}{BatStyler} & SD ($\downarrow$) & 0.07 & 0.12 & 0.13 & 0.04 & 0.09 & 0.17 & 0.04 & 0.12 & 0.27 & 0.03 & 0.1 & 0.06 \\
				& SC ($\uparrow$) & 0.11 & 0.29 & 0.61 & 0.07 & 0.31 & 0.59 & 0.09 & 0.16 & 0.31 & 0.13 & 0.37 & 0.59 \\    
				\bottomrule
			\end{tabular}
			\label{tab:lamda}
		}
	\end{table*}%
	
	\begin{figure*}[t]
		\centering
		\includegraphics[width=18cm, height=3.2cm]{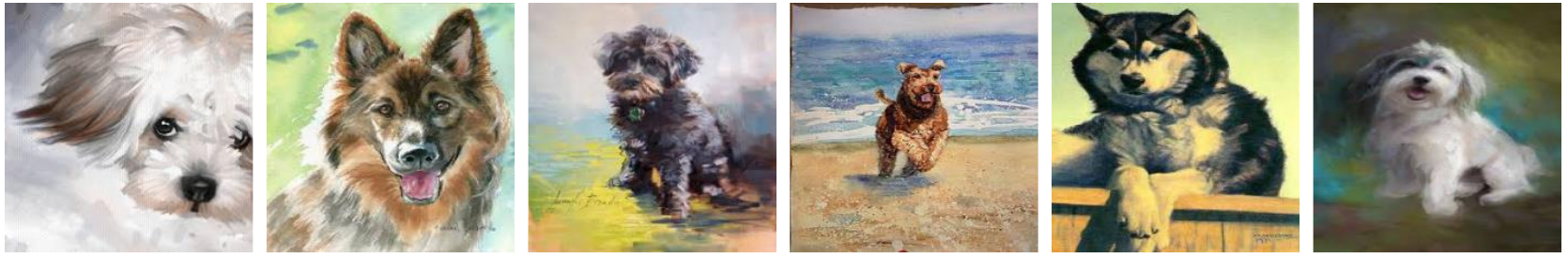}
		\caption{Text-to-Image synthesis results using well-learned pseudo style words. We randomly select 6 style-content prompt, we set content as ``dog'', and retrieve images from PACS, the retrieved images and these six style-content prompt have a average cosine similarity over 0.81, which implies a high semantic similarity that pseudo-styles retain. } 
		\label{figure:show}
	\end{figure*}
	
	\begin{figure*}[t]
		\centering
		{\includegraphics[width=0.3\linewidth]{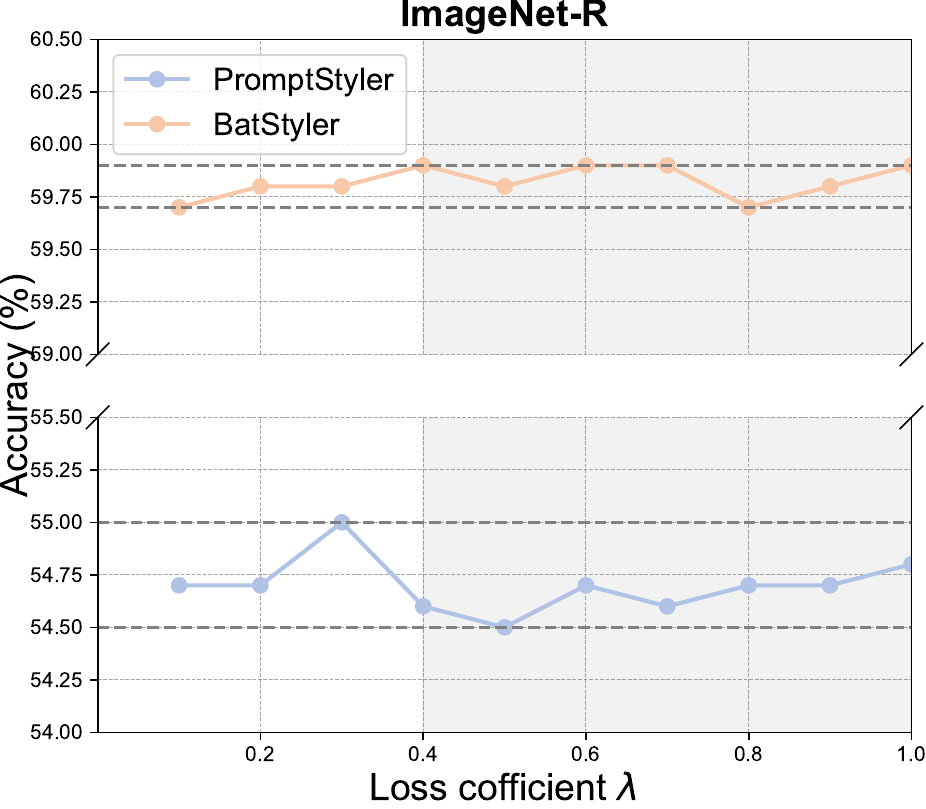}}
		{\includegraphics[width=0.3\linewidth]{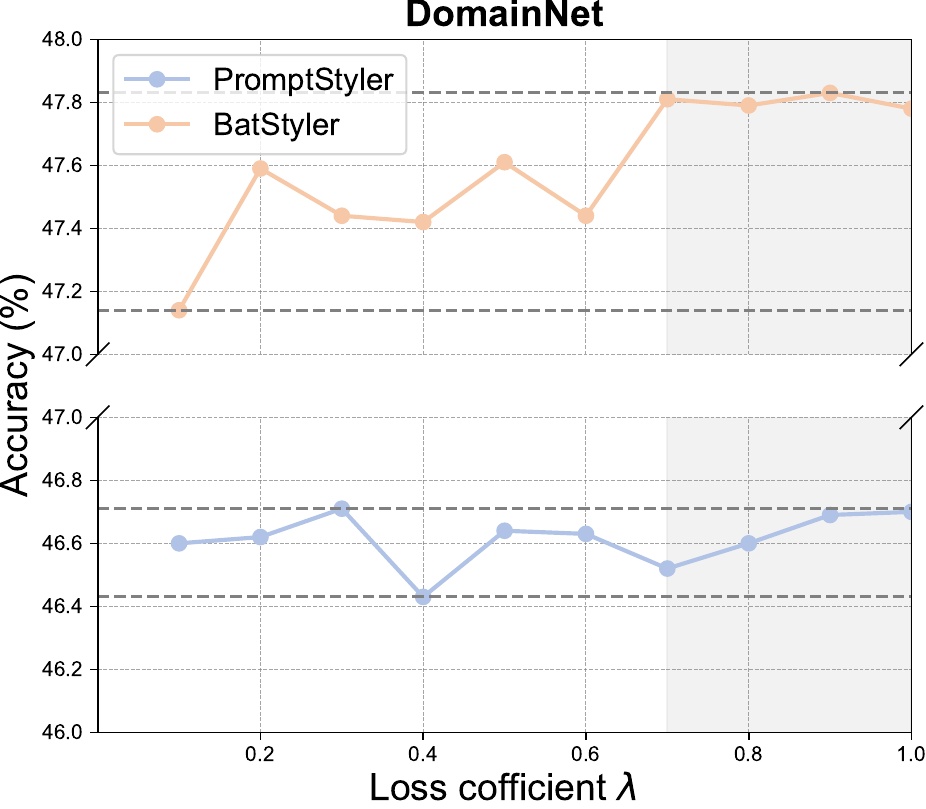}}
		{\includegraphics[width=0.3\linewidth]{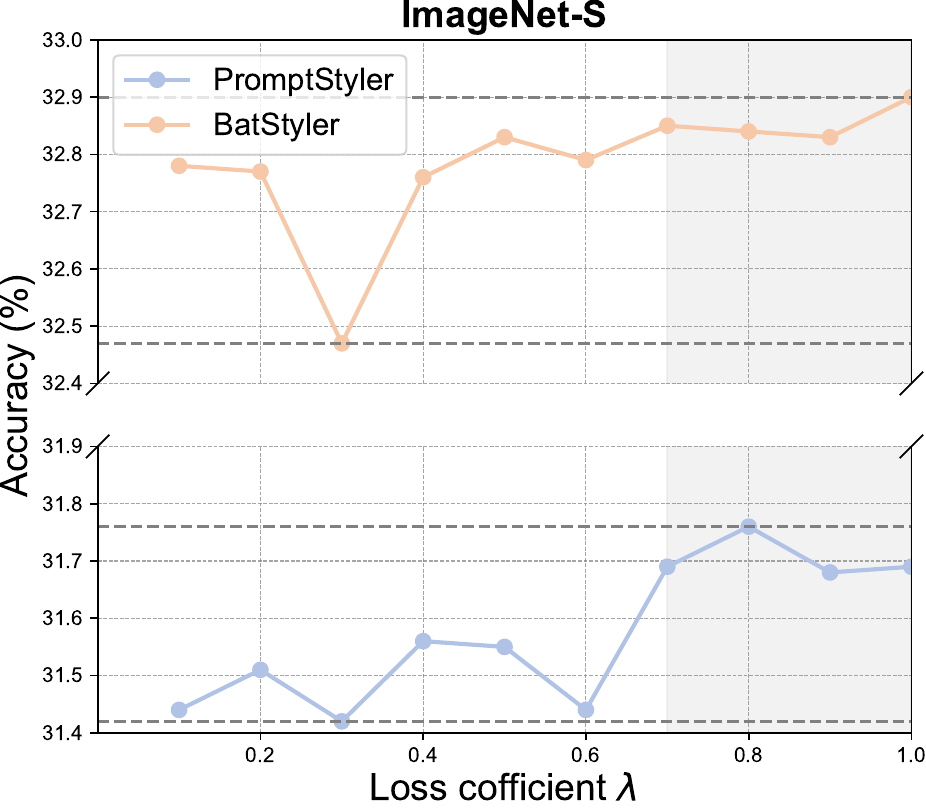}}
		\caption{Performance on different $\lambda$ across multi-category datasets. When $\lambda$ is around 0.8, there is no significant fluctuations on our method in gray area. }
		\label{figure:md}
	\end{figure*}
	
	\textbf{Comparison under different classifier initialization.} We compare neural collapse with random initialization but not trainable, and a learnable classifier in Tab. \ref{table:classifier}. According to the comparison, we find that the proposed NC-initialization can obtain better performance than the random classifier and trainable classifier, which thanks to the fact that the neural collapse framework is a optimal classifier structure. It can generate boundary vectors with the maximum margin, thereby representing these vectors as a classifier, ensuring the superiority and distinctiveness of the templates.
	
	\subsection{Further Analysis}
	\label{sec:EXP-FA}
	
	\textbf{Analysis of Coarse Semantic Generation.} In this section, we investigate the impact of different numbers of coarse-grained semantics of each cluster \textit{C}, as shown in Tab. \ref{tab:N}. Due to the comparable performance, we only evaluate the impact on multi-category datasets. As observed in figure, in ImageNet-R, ImageNet-S and DomainNet, after coarse-grained extraction, the performance is fluctuating obviously, especially in ImageNet-R. The motivation of proposed method is to reduce the number of categories \textit{N} in multi-category scenarios to enhance style diversity. If too many semantics are extracted, it will again lead to poor style diversity. Therefore, it is necessary to limit the number of semantics within an appropriate range. It can be observed that when \textit{C} is relatively large, the performance significantly declines. When extracting one coarse-grained semantic, the extracted semantic may not be sufficiently informative to fully represent the cluster it stands for, hence the need to extract multiple semantics. Additionally, between 3 and 5, the model shows no significant fluctuations, so we set \textit{C} to 3, which achieves good performance with fewer semantic costs.
	
	\textbf{Analysis of Text-to-Image synthesis.} In Fig. \ref{figure:show}, we select 6 different style-content prompts randomly and retrieve images from PACS, the retrieved images and these six style-content prompt have a average cosine similarity over 0.81, which implies a enough semantic that pseudo-styles retain. It is an evidence that proposed method injects enough semantics into pseudo-styles and we achieve a better diversity.

	\textbf{Analysis of Loss coefficient.} In Sec. \ref{sec:dis}, we have discussed the impact of $\lambda$ on style diversity. In this part, we provide a detailed explanation of it. The format of adding loss coefficient is formulated in Eq. \ref{eq:7}. The detailed results are shown in Tab. \ref{tab:lamda}. The lower SD, the better style diversity; the higher SC, the better semantic consistency. To further validate the efficacy of proposed method, we also conduct experiments on BatStyler. Note that the loss of our method do not consist of $\lambda$, here, it is merely for the sake of comparison. From the table, it is worth noting that our method has achieved enhanced diversity learning and semantic consistency learning. Style diversity increases with the decrease of $\lambda$, but semantic consistency also weakens, which is not what we desire. In our method, this phenomenon also occurs, so it is not feasible to weaken semantic consistency constraints by simply decreasing $\lambda$. BatStyler demonstrates better style diversity and semantic consistency under the same conditions compared to PromptStyler, proving the effectiveness of proposed approach.
	

	To validate the stability of our method, we conduct another experiment, as shown in Fig. \ref{figure:md}. We only evaluate performance on multi-category datasets. The model's performance fluctuates slightly and is not sensitive to $\lambda$, especially in gray area. We think that there is no need to expend resources to adjust the
	hyperparameter $\lambda$ for specific tasks. Given the slight performance fluctuations and to avoid the complex
	search for the optimal $\lambda$, we directly set $\lambda$ to 1, so $\lambda$ does not appear in the loss function Eq. \ref{formula:1}.

	\textbf{The number of pseudo-styles.} We conduct experiments with the number of pseudo-styles at [80, 200, 400, 600, 1000], and it can be observed that performance does not consistently increase with the number of styles increase, but there is a general upward trend in ImageNet-R and ImageNet-S. In theory, the more pseudo-styles there are, the lower the density between styles, which does not affect the overall diversity, and thus will not bring a continuous improvement. In our experiments, the default number of pseudo-styles is set to 80 for ease of comparison with other methods. 
	
	\begin{figure}[t]
		\centering
		\includegraphics[width=0.7\linewidth]{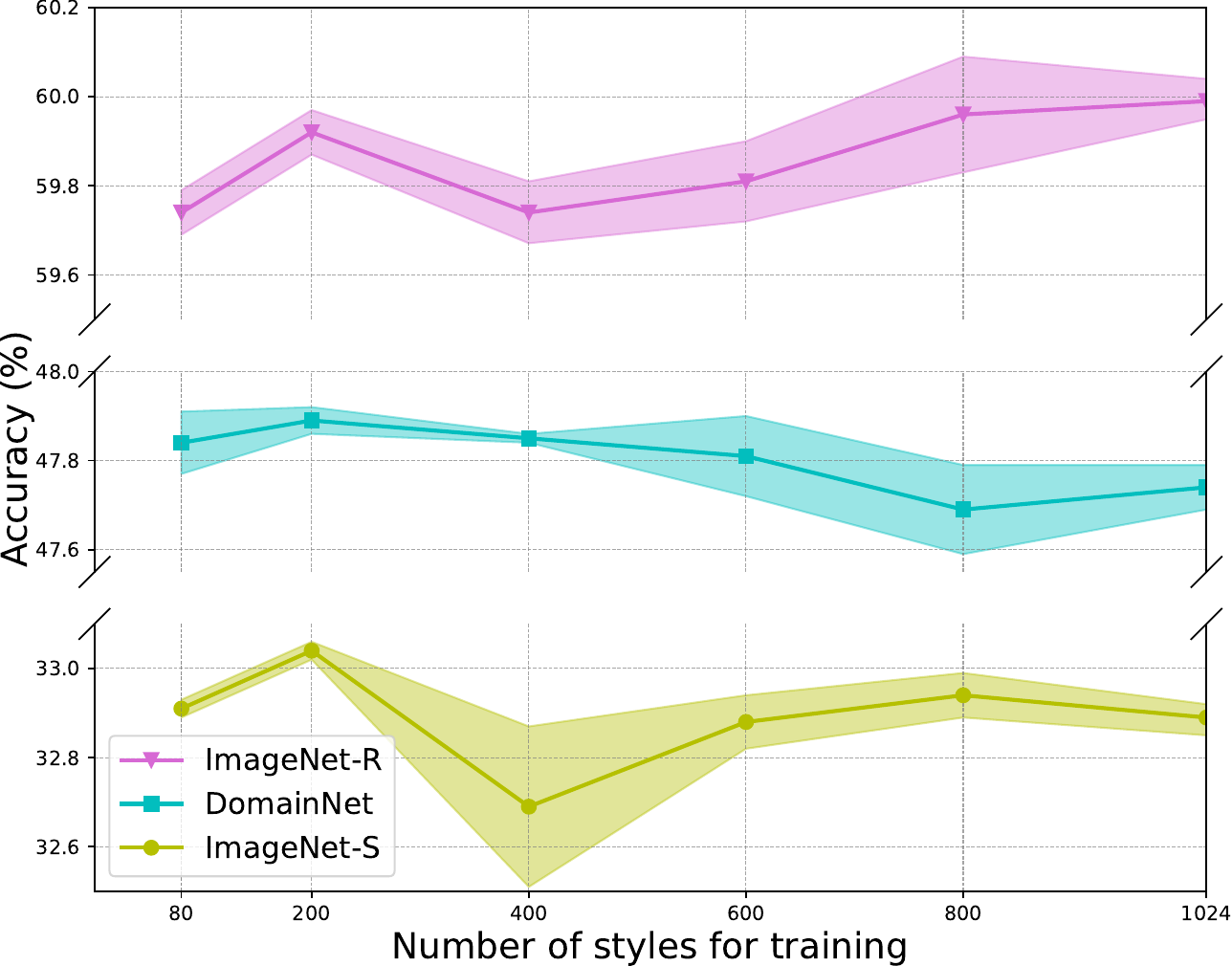}
		\caption{Performance on different number of styles across ImageNet-R, DomainNet and ImageNet-S. All experiments are implemented on CLIP's RN50 model. } 
		\label{figure:ns}
	\end{figure}
	
	\textbf{Analysis of computational resources scalability on large-scale datasets.} The currently existing classification datasets are limited in scale, and the number of categories in downstream tasks is only a small subset in real-world scenarios. In source-free domain generalization, only the category names are accessible, so there is no need to consider the impact of image data scale on the model. When the number of downstream categories is so large that there is no common semantics, we can repeatedly perform binary clustering until meaningful semantic information is extracted. By using multi-level clustering to transform a large semantic set into several relatively smaller coarse-grained semantic sets, we can achieve the scalability of computational resources on large-scale datasets in real world.
	
	
	\section{Conclusion}
	\label{section:conclusion}
	In this paper, we have proposed an effective strategy to synthesize pseudo-styles, called BatStyler, which utilized to enhance the style diversity in multi-category scenarios without compromising the semantic consistency, and it realizes parallel training. Although we have achieved good performance, since we build upon the CLIP, the performance heavily relies on CLIP's joint space of vision and language, which implies if the vision and language is not well aligned, the performance may also suffer from deterioration.
	
	\ifCLASSOPTIONcaptionsoff
	\newpage
	\fi

	\ifCLASSOPTIONcaptionsoff
	\newpage
	\fi
	\bibliographystyle{IEEEtran}
	\bibliography{sigproc}

\begin{thebibliography}{10}
\providecommand{\url}[1]{#1}
\csname url@samestyle\endcsname
\providecommand{\newblock}{\relax}
\providecommand{\bibinfo}[2]{#2}
\providecommand{\BIBentrySTDinterwordspacing}{\spaceskip=0pt\relax}
\providecommand{\BIBentryALTinterwordstretchfactor}{4}
\providecommand{\BIBentryALTinterwordspacing}{\spaceskip=\fontdimen2\font plus
\BIBentryALTinterwordstretchfactor\fontdimen3\font minus
  \fontdimen4\font\relax}
\providecommand{\BIBforeignlanguage}[2]{{%
\expandafter\ifx\csname l@#1\endcsname\relax
\typeout{** WARNING: IEEEtran.bst: No hyphenation pattern has been}%
\typeout{** loaded for the language `#1'. Using the pattern for}%
\typeout{** the default language instead.}%
\else
\language=\csname l@#1\endcsname
\fi
#2}}
\providecommand{\BIBdecl}{\relax}
\BIBdecl

\bibitem{zhu2024solid}
J.~Zhu, X.~Liu, Z.~Liu, Y.~Deng, J.~Xu, K.~Liu, R.~Zhang, X.~Meng, P.~Fei,
  T.~Yu \emph{et~al.}, ``Solid: minimizing tissue distortion for brain-wide
  profiling of diverse architectures,'' \emph{Nature Communications}, p. 8303,
  2024.

\bibitem{BalakrishnanZSG19}
G.~Balakrishnan, A.~Zhao, M.~R. Sabuncu, J.~V. Guttag, and A.~V. Dalca,
  ``Voxelmorph: {A} learning framework for deformable medical image
  registration,'' \emph{IEEE Transactions on Medical Imaging}, pp. 1788--1800,
  2019.

\bibitem{he2016deep}
K.~He, X.~Zhang, S.~Ren, and J.~Sun, ``Deep residual learning for image
  recognition,'' in \emph{Conference on Computer Vision and Pattern Recognition
  (CVPR)}, 2016, pp. 770--778.

\bibitem{krizhevsky2012imagenet}
A.~Krizhevsky, I.~Sutskever, and G.~E. Hinton, ``Imagenet classification with
  deep convolutional neural networks,'' in \emph{Advances in Neural Information
  Processing Systems (NeurIPS)}, 2012, pp. 1106--1114.

\bibitem{simonyan2014very}
K.~Simonyan and A.~Zisserman, ``Very deep convolutional networks for
  large-scale image recognition,'' in \emph{International Conference on
  Learning Representations (ICLR)}, 2014.

\bibitem{Peng23}
S.~Peng, ``Application of medical image detection technology based on deep
  learning in pneumoconiosis diagnosis,'' \emph{Data Intelligence}, vol.~5, pp.
  1033--1047, 2023.

\bibitem{huang2017densely}
G.~Huang, Z.~Liu, L.~Van Der~Maaten, and K.~Q. Weinberger, ``Densely connected
  convolutional networks,'' in \emph{Conference on Computer Vision and Pattern
  Recognition (CVPR)}, 2017, pp. 2261--2269.

\bibitem{wang2017residual}
F.~Wang, M.~Jiang, C.~Qian, S.~Yang, C.~Li, H.~Zhang, X.~Wang, and X.~Tang,
  ``Residual attention network for image classification,'' in \emph{Conference
  on Computer Vision and Pattern Recognition (CVPR)}, 2017, pp. 6450--6458.

\bibitem{carion2020end}
N.~Carion, F.~Massa, G.~Synnaeve, N.~Usunier, A.~Kirillov, and S.~Zagoruyko,
  ``End-to-end object detection with transformers,'' in \emph{European
  Conference on Computer Vision (ECCV)}, vol. 12346, 2020, pp. 213--229.

\bibitem{faster2015towards}
S.~Ren, K.~He, R.~B. Girshick, and J.~Sun, ``Faster {R-CNN:} towards real-time
  object detection with region proposal networks,'' in \emph{Advances in Neural
  Information Processing Systems (NeurIPS)}, 2015, pp. 91--99.

\bibitem{hoffman2013one}
J.~Hoffman, E.~Tzeng, J.~Donahue, Y.~Jia, K.~Saenko, and T.~Darrell, ``One-shot
  adaptation of supervised deep convolutional models,'' in \emph{International
  Conference on Learning Representations (ICLR)}, 2014.

\bibitem{lee2022fifo}
S.~Lee, T.~Son, and S.~Kwak, ``Fifo: Learning fog-invariant features for foggy
  scene segmentation,'' in \emph{Conference on Computer Vision and Pattern
  Recognition (CVPR)}, 2022, pp. 18\,889--18\,899.

\bibitem{saito2019semi}
K.~Saito, D.~Kim, S.~Sclaroff, T.~Darrell, and K.~Saenko, ``Semi-supervised
  domain adaptation via minimax entropy,'' in \emph{International Conference on
  Computer Vision (ICCV)}, 2019, pp. 8049--8057.

\bibitem{LiuXLTZ23}
Y.~Liu, Z.~Xiong, Y.~Li, X.~Tian, and Z.~Zha, ``Domain generalization via
  encoding and resampling in a unified latent space,'' \emph{IEEE Transactions
  on Multimedia (TMM)}, vol.~25, pp. 126--139, 2023.

\bibitem{JinLZC22}
X.~Jin, C.~Lan, W.~Zeng, and Z.~Chen, ``Style normalization and restitution for
  domain generalization and adaptation,'' \emph{IEEE Transactions on Multimedia
  (TMM)}, vol.~24, pp. 3636--3651, 2022.

\bibitem{QiWSG23}
L.~Qi, L.~Wang, Y.~Shi, and X.~Geng, ``A novel mix-normalization method for
  generalizable multi-source person re-identification,'' \emph{IEEE
  Transactions on Multimedia (TMM)}, vol.~25, pp. 4856--4867, 2023.

\bibitem{ChenLLT23}
H.~Chen, Y.~Lin, B.~Li, and S.~Tan, ``Learning features of intra-consistency
  and inter-diversity: Keys toward generalizable deepfake detection,''
  \emph{IEEE Transactions on Circuits and Systems for Video Technology
  (TCSVT)}, vol.~33, pp. 1468--1480, 2023.

\bibitem{Cho_2023_ICCV}
J.~Cho, G.~Nam, S.~Kim, H.~Yang, and S.~Kwak, ``Promptstyler: Prompt-driven
  style generation for source-free domain generalization,'' in
  \emph{International Conference on Computer Vision (ICCV)}, 2023, pp.
  15\,656--15\,666.

\bibitem{0002CKRT22}
A.~Frikha, H.~Chen, D.~Krompa{\ss}, T.~A. Runkler, and V.~Tresp, ``Towards
  data-free domain generalization,'' in \emph{ACML}, 2022, pp. 327--342.

\bibitem{niu2022domain}
H.~Niu, H.~Li, F.~Zhao, and B.~Li, ``Domain-unified prompt representations for
  source-free domain generalization,'' \emph{arXiv preprint arXiv:2209.14926},
  2022.

\bibitem{KunduKSJB21}
J.~N. Kundu, A.~R. Kulkarni, A.~Singh, V.~Jampani, and R.~V. Babu, ``Generalize
  then adapt: Source-free domain adaptive semantic segmentation,'' in
  \emph{ICCV}, 2021, pp. 7026--7036.

\bibitem{fang2022source}
Y.~Fang, P.~Yap, W.~Lin, H.~Zhu, and M.~Liu, ``Source-free unsupervised domain
  adaptation: {A} survey,'' \emph{Neural Networks}, vol. 174, p. 106230, 2024.

\bibitem{huang2021model}
J.~Huang, D.~Guan, A.~Xiao, and S.~Lu, ``Model adaptation: Historical
  contrastive learning for unsupervised domain adaptation without source
  data,'' in \emph{Advances in Neural Information Processing Systems
  (NeurIPS)}, 2021, pp. 3635--3649.

\bibitem{0001S0Z24}
S.~Tang, W.~Su, M.~Ye, and X.~Zhu, ``Source-free domain adaptation with frozen
  multimodal foundation model,'' in \emph{CVPR}, 2024, pp. 23\,711--23\,720.

\bibitem{radford2021learning}
A.~Radford, J.~W. Kim, C.~Hallacy, A.~Ramesh, G.~Goh, S.~Agarwal, G.~Sastry,
  A.~Askell, P.~Mishkin, J.~Clark \emph{et~al.}, ``Learning transferable visual
  models from natural language supervision,'' in \emph{International Conference
  on Machine Learning (ICML)}, 2021, pp. 8748--8763.

\bibitem{FanWKYGZ21}
X.~Fan, Q.~Wang, J.~Ke, F.~Yang, B.~Gong, and M.~Zhou, ``Adversarially adaptive
  normalization for single domain generalization,'' in \emph{Conference on
  Computer Vision and Pattern Recognition (CVPR)}, 2021, pp. 8208--8217.

\bibitem{QiaoZP20}
F.~Qiao, L.~Zhao, and X.~Peng, ``Learning to learn single domain
  generalization,'' in \emph{Conference on Computer Vision and Pattern
  Recognition (CVPR)}, 2020, pp. 12\,553--12\,562.

\bibitem{ZhangMLHT18}
L.~Zhang, B.~Ma, G.~Li, Q.~Huang, and Q.~Tian, ``Generalized semi-supervised
  and structured subspace learning for cross-modal retrieval,'' \emph{IEEE
  Transactions on Multimedia (TMM)}, vol.~20, no.~1, pp. 128--141, 2018.

\bibitem{FangOLY23}
W.~Fang, C.~Ouyang, Q.~Lin, and Y.~Yuan, ``Three heads better than one: Pure
  entity, relation label and adversarial training for cross-domain few-shot
  relation extraction,'' \emph{Data Intelligence}, vol.~5, pp. 807--823, 2023.

\bibitem{NamLPYY21}
H.~Nam, H.~Lee, J.~Park, W.~Yoon, and D.~Yoo, ``Reducing domain gap by reducing
  style bias,'' in \emph{Conference on Computer Vision and Pattern Recognition
  (CVPR)}, 2021, pp. 8690--8699.

\bibitem{NurielBW21}
O.~Nuriel, S.~Benaim, and L.~Wolf, ``Permuted adain: Reducing the bias towards
  global statistics in image classification,'' in \emph{Conference on Computer
  Vision and Pattern Recognition (CVPR)}, 2021, pp. 9482--9491.

\bibitem{zhou2020domain}
F.~Zhou, Z.~Jiang, C.~Shui, B.~Wang, and B.~Chaib-draa, ``Domain generalization
  with optimal transport and metric learning,'' \emph{arXiv preprint
  arXiv:2007.10573}, 2020.

\bibitem{PengBXHSW19}
X.~Peng, Q.~Bai, X.~Xia, Z.~Huang, K.~Saenko, and B.~Wang, ``Moment matching
  for multi-source domain adaptation,'' in \emph{International Conference on
  Computer Vision (ICCV)}, 2019, pp. 1406--1415.

\bibitem{ZhaoXCCLZXKPSXY18}
J.~Zhao, L.~Xiong, Y.~Cheng, Y.~Cheng, J.~Li, L.~Zhou, Y.~Xu, J.~Karlekar,
  S.~Pranata, S.~Shen, J.~Xing, S.~Yan, and J.~Feng, ``3d-aided deep
  pose-invariant face recognition,'' in \emph{IJCAI}, 2018, pp. 1184--1190.

\bibitem{PanLST18}
X.~Pan, P.~Luo, J.~Shi, and X.~Tang, ``Two at once: Enhancing learning and
  generalization capacities via ibn-net,'' in \emph{European Conference on
  Computer Vision (ECCV)}, 2018, pp. 484--500.

\bibitem{JiaRH19}
J.~Jia, Q.~Ruan, and T.~M. Hospedales, ``Frustratingly easy person
  re-identification: Generalizing person re-id in practice,'' in \emph{British
  Machine Vision Conference (BMVC)}, 2019, p. 117.

\bibitem{ChenZSYWTY23}
D.~Chen, Y.~Zhuang, Z.~Shen, C.~Yang, G.~Wang, S.~Tang, and Y.~Yang,
  ``Cross-modal data augmentation for tasks of different modalities,''
  \emph{IEEE Transactions on Multimedia (TMM)}, vol.~25, pp. 7814--7824, 2023.

\bibitem{ZhaoXKLZWPSYF17}
J.~Zhao, L.~Xiong, J.~Karlekar, J.~Li, F.~Zhao, Z.~Wang, S.~Pranata, S.~Shen,
  S.~Yan, and J.~Feng, ``Dual-agent gans for photorealistic and identity
  preserving profile face synthesis,'' in \emph{NeurIPS}, 2017, pp. 66--76.

\bibitem{ZhaoCC00LLYF19}
J.~Zhao, Y.~Cheng, Y.~Cheng, Y.~Yang, F.~Zhao, J.~Li, H.~Liu, S.~Yan, and
  J.~Feng, ``Look across elapse: Disentangled representation learning and
  photorealistic cross-age face synthesis for age-invariant face recognition,''
  in \emph{AAAI}, 2019, pp. 9251--9258.

\bibitem{JiaYXCPPLSLD21}
C.~Jia, Y.~Yang, Y.~Xia, Y.~Chen, Z.~Parekh, H.~Pham, Q.~V. Le, Y.~Sung, Z.~Li,
  and T.~Duerig, ``Scaling up visual and vision-language representation
  learning with noisy text supervision,'' in \emph{ICML}, 2021, pp. 4904--4916.

\bibitem{0001LXH22}
J.~Li, D.~Li, C.~Xiong, and S.~C.~H. Hoi, ``{BLIP:} bootstrapping
  language-image pre-training for unified vision-language understanding and
  generation,'' in \emph{International Conference on Machine Learning (ICML)},
  2022, pp. 12\,888--12\,900.

\bibitem{zhu2023minigpt}
D.~Zhu, J.~Chen, X.~Shen, X.~Li, and M.~Elhoseiny, ``Minigpt-4: Enhancing
  vision-language understanding with advanced large language models,''
  \emph{arXiv preprint arXiv:2304.10592}, 2023.

\bibitem{ZhengWWQB22}
Q.~Zheng, H.~Wen, M.~Wang, G.~Qi, and C.~Bai, ``Faster zero-shot multi-modal
  entity linking via visual-linguistic representation,'' \emph{Data
  Intelligence}, vol.~4, pp. 493--508, 2022.

\bibitem{ZhangHHW0Y23}
Y.~Zhang, J.~Z. HaoChen, S.~Huang, K.~Wang, J.~Zou, and S.~Yeung, ``Diagnosing
  and rectifying vision models using language,'' in \emph{International
  Conference on Learning Representations (ICLR)}, 2023.

\bibitem{papyan2020prevalence}
V.~Papyan, X.~Han, and D.~L. Donoho, ``Prevalence of neural collapse during the
  terminal phase of deep learning training,'' \emph{Proceedings of the National
  Academy of Sciences (PNAS)}, 2020.

\bibitem{YangCLXLT22}
Y.~Yang, S.~Chen, X.~Li, L.~Xie, Z.~Lin, and D.~Tao, ``Inducing neural collapse
  in imbalanced learning: Do we really need a learnable classifier at the end
  of deep neural network?'' in \emph{Advances in Neural Information Processing
  Systems (NeurIPS)}, 2022.

\bibitem{XieYCH23}
L.~Xie, Y.~Yang, D.~Cai, and X.~He, ``Neural collapse inspired
  attraction-repulsion-balanced loss for imbalanced learning,''
  \emph{Neurocomputing}, vol. 527, pp. 60--70, 2023.

\bibitem{XiaoFTZLCW24}
R.~Xiao, L.~Feng, K.~Tang, J.~Zhao, Y.~Li, G.~Chen, and H.~Wang, ``Targeted
  representation alignment for open-world semi-supervised learning,'' in
  \emph{CVPR}, 2024, pp. 23\,072--23\,082.

\bibitem{HuWNTN24}
Z.~Hu, Y.~Wang, H.~Ning, Y.~Tai, and F.~Nie, ``Neural collapse inspired
  semi-supervised learning with fixed classifier,'' \emph{Information
  Sciences}, vol. 667, p. 120469, 2024.

\bibitem{RanLLTNT24}
H.~Ran, W.~Li, L.~Li, S.~Tian, X.~Ning, and P.~Tiwari, ``Learning optimal
  inter-class margin adaptively for few-shot class-incremental learning via
  neural collapse-based meta-learning,'' \emph{Information Processing and
  Management}, p. 103664, 2024.

\bibitem{YangYLLTT23}
Y.~Yang, H.~Yuan, X.~Li, Z.~Lin, P.~H.~S. Torr, and D.~Tao, ``Neural collapse
  inspired feature-classifier alignment for few-shot class-incremental
  learning,'' in \emph{International Conference on Learning Representations
  (ICLR)}, 2023.

\bibitem{Zhu0ZYLKW24}
D.~Zhu, Z.~Li, M.~Zhang, J.~Yuan, J.~Liu, K.~Kuang, and C.~Wu, ``Neural
  collapse anchored prompt tuning for generalizable vision-language models,''
  in \emph{KDD}, 2024, pp. 4631--4640.

\bibitem{zhu2023bridging}
D.~Zhu, Y.~Li, M.~Zhang, J.~Yuan, J.~Liu, K.~Kuang, and C.~Wu, ``Bridging the
  gap: neural collapse inspired prompt tuning for generalization under class
  imbalance,'' \emph{arXiv preprint arXiv:2306.15955}, 2023.

\bibitem{ArthurV07}
D.~Arthur and S.~Vassilvitskii, ``k-means++: the advantages of careful
  seeding,'' in \emph{Symposium on Discrete Algorithms (ACM-SIAM)}, 2007, pp.
  1027--1035.

\bibitem{achiam2023gpt}
J.~Achiam, S.~Adler, S.~Agarwal, L.~Ahmad, I.~Akkaya, F.~L. Aleman, D.~Almeida,
  J.~Altenschmidt, S.~Altman, S.~Anadkat \emph{et~al.}, ``Gpt-4 technical
  report,'' \emph{arXiv preprint arXiv:2303.08774}, 2023.

\bibitem{DengGXZ19}
J.~Deng, J.~Guo, N.~Xue, and S.~Zafeiriou, ``Arcface: Additive angular margin
  loss for deep face recognition,'' in \emph{Conference on Computer Vision and
  Pattern Recognition (CVPR)}, 2019, pp. 4690--4699.

\bibitem{BeleBBJRB24}
P.~Bele, V.~Bundele, A.~Bhattacharya, A.~Jha, G.~Roig, and B.~Banerjee,
  ``Learning class and domain augmentations for single-source open-domain
  generalization,'' in \emph{WACV}, 2024, pp. 1805--1815.

\bibitem{LiLWLT23}
J.~Li, Y.~Li, H.~Wang, C.~Liu, and J.~Tan, ``Exploring explicitly disentangled
  features for domain generalization,'' \emph{IEEE Transactions on Circuits and
  Systems for Video Technology (TCSVT)}, vol.~33, pp. 6360--6373, 2023.

\bibitem{QiYSG24}
L.~Qi, H.~Yang, Y.~Shi, and X.~Geng, ``Normaug: Normalization-guided
  augmentation for domain generalization,'' \emph{IEEE Transactions on Image
  Processing (TIP)}, pp. 1419--1431, 2024.

\bibitem{ChenWZSMD24}
Z.~Chen, W.~Wang, Z.~Zhao, F.~Su, A.~Men, and Y.~Dong, ``Instance paradigm
  contrastive learning for domain generalization,'' \emph{IEEE Transactions on
  Circuits and Systems for Video Technology (TCSVT)}, vol.~34, pp. 1032--1042,
  2024.

\bibitem{HeWTL24}
J.~He, L.~Wu, C.~Tao, and F.~Lv, ``Source-free domain adaptation with
  unrestricted source hypothesis,'' \emph{Pattern Recognit.}, vol. 149, p.
  110246, 2024.

\bibitem{ZhangWSZLZ24}
L.~Zhang, Y.~Wang, R.~Song, M.~Zhang, X.~Li, and W.~Zhang, ``Neighborhood-aware
  mutual information maximization for source-free domain adaptation,''
  \emph{IEEE Transactions on Multimedia (TMM)}, vol.~26, pp. 9564--9574, 2024.

\bibitem{RothKKVSA23}
K.~Roth, J.~Kim, A.~S. Koepke, O.~Vinyals, C.~Schmid, and Z.~Akata, ``Waffling
  around for performance: Visual classification with random words and broad
  concepts,'' in \emph{International Conference on Computer Vision (ICCV)},
  2023, pp. 15\,700--15\,711.

\bibitem{BoseJFS0B24}
S.~Bose, A.~Jha, E.~Fini, M.~Singha, E.~Ricci, and B.~Banerjee, ``Stylip:
  Multi-scale style-conditioned prompt learning for clip-based domain
  generalization,'' in \emph{WACV}, 2024, pp. 5530--5540.

\bibitem{zhang2024promptta}
H.~Zhang, S.~Bai, W.~Zhou, J.~Fu, and B.~Chen, ``Promptta: Prompt-driven text
  adapter for source-free domain generalization,'' \emph{arXiv preprint
  arXiv:2409.14163}, 2024.

\bibitem{tang2024dpstyler}
Y.~Tang, Y.~Wan, L.~Qi, and X.~Geng, ``Dpstyler: Dynamic promptstyler for
  source-free domain generalization,'' \emph{arXiv preprint arXiv:2403.16697},
  2024.

\bibitem{MenonV23}
S.~Menon and C.~Vondrick, ``Visual classification via description from large
  language models,'' in \emph{ICLR}, 2023.

\bibitem{RenS023}
Z.~Ren, Y.~Su, and X.~Liu, ``Chatgpt-powered hierarchical comparisons for image
  classification,'' in \emph{NeurIPS}, A.~Oh, T.~Naumann, A.~Globerson,
  K.~Saenko, M.~Hardt, and S.~Levine, Eds., 2023.

\bibitem{LiYSH17}
D.~Li, Y.~Yang, Y.~Song, and T.~M. Hospedales, ``Deeper, broader and artier
  domain generalization,'' in \emph{International Conference on Computer Vision
  (ICCV)}, 2017, pp. 5543--5551.

\bibitem{FangXR13}
C.~Fang, Y.~Xu, and D.~N. Rockmore, ``Unbiased metric learning: On the
  utilization of multiple datasets and web images for softening bias,'' in
  \emph{International Conference on Computer Vision (ICCV)}, 2013, pp.
  1657--1664.

\bibitem{VenkateswaraECP17}
H.~Venkateswara, J.~Eusebio, S.~Chakraborty, and S.~Panchanathan, ``Deep
  hashing network for unsupervised domain adaptation,'' in \emph{Conference on
  Computer Vision and Pattern Recognition (CVPR)}, 2017, pp. 5385--5394.

\bibitem{HendrycksBMKWDD21}
D.~Hendrycks, S.~Basart, N.~Mu, S.~Kadavath, F.~Wang, E.~Dorundo, R.~Desai,
  T.~Zhu, S.~Parajuli, M.~Guo, D.~Song, J.~Steinhardt, and J.~Gilmer, ``The
  many faces of robustness: {A} critical analysis of out-of-distribution
  generalization,'' in \emph{International Conference on Computer Vision
  (ICCV)}, 2021, pp. 8320--8329.

\bibitem{GaoLYCHT23}
S.~Gao, Z.~Li, M.~Yang, M.~Cheng, J.~Han, and P.~H.~S. Torr, ``Large-scale
  unsupervised semantic segmentation,'' \emph{IEEE Transactions on Pattern
  Analysis and Machine Intelligence Information (TPAMI)}, vol.~45, no.~6, pp.
  7457--7476, 2023.

\bibitem{van2008visualizing}
L.~Van~der Maaten and G.~Hinton, ``Visualizing data using t-sne,''
  \emph{Journal of Machine Learning Research (JMLR)}, vol.~9, no.~11, 2008.

\end{thebibliography}
	\vspace{-0.7cm}

	\vfill
	

\end{document}